\newif\ifJOURNAL
\JOURNALfalse
\newif\ifarXiv
\arXivfalse
\newif\ifWP
\WPfalse
\newif\ifFULL
\FULLfalse
\newif\ifLATIN
\LATINfalse

\arXivtrue

\ifarXiv\LATINtrue\fi	

\newif\ifnotJOURNAL	
\notJOURNALtrue
\ifJOURNAL\notJOURNALfalse\fi

\newif\ifnotarXiv	
\notarXivtrue
\ifarXiv\notarXivfalse\fi

\newif\ifTR		
\TRfalse
\ifarXiv\TRtrue\fi
\ifWP\TRtrue\fi
\newif\ifnotTR
\notTRtrue
\ifarXiv\notTRfalse\fi
\ifWP\notTRfalse\fi

\newif\ifnotLATIN	
\notLATINtrue
\ifLATIN\notLATINfalse\fi

\ifJOURNAL
  \newcommand{\GTPVII}{vovk/shafer:2005RSS}
  \newcommand{\GTPVIII}{vovk/etal:2005AIStatslocal}
  \newcommand{\GTPX}{vovk/etal:2005ALT}
  
  \newcommand{\GTPXIII}{vovk:2005ALT-GTP13}
  \newcommand{\GTPXIV}{vovk:2005ALT-GTP14}
  
\fi
\ifarXiv
  \newcommand{\GTPVII}{GTP7}
  \newcommand{\GTPVIII}{GTP8arXiv}
  \newcommand{\GTPX}{GTP10arXiv}
  
  \newcommand{\GTPXIII}{GTP13arXiv}
  \newcommand{\GTPXIV}{GTP14arXiv}
\fi
\ifWP
  \newcommand{\GTPVII}{GTP7}
  \newcommand{\GTPVIII}{GTP8}
  \newcommand{\GTPX}{GTP10}
  
  \newcommand{\GTPXIII}{GTP13}
  \newcommand{\GTPXIV}{GTP14}
\fi
\ifFULL
  \newcommand{\GTPVII}{GTP7}
  \newcommand{\GTPVIII}{GTP8arXiv}
  \newcommand{\GTPX}{GTP10arXiv}
  
  \newcommand{\GTPXIII}{GTP13arXiv}
  \newcommand{\GTPXIV}{GTP14arXiv}
\fi

\ifnotLATIN
  \newcommand{\Levin}{levin:1976uniform}
  \newcommand{\Takemura}{takemura:2004}
\fi
\ifLATIN
  \newcommand{\Levin}{levin:1976uniformlatin}
  \newcommand{\Takemura}{takemura:2004latin}
\fi

\ifJOURNAL
\documentclass[toc]{article}
\usepackage{amsmath,amsfonts,amssymb,latexsym,graphicx}
\newcommand{\Extra}[1]{}
\fi

\ifarXiv
\documentclass[toc]{article}
\usepackage{amsmath,amsfonts,amssymb,latexsym,graphicx}
\newcommand{\Extra}[1]{}
\fi

\ifWP
\documentclass[toc]{gtarticle}
\usepackage{amsmath,amsfonts,amssymb,latexsym,epsfig,graphicx}
\renewcommand{\Extra}[1]{#1}
\fi

\ifFULL
\documentclass{article}
\usepackage{amsmath,amsfonts,amssymb,latexsym,color,epigraph,graphicx}
\newcommand{\Extra}[1]{\red{#1}}
\newcommand{\red}[1]{\textcolor{red}{#1}}

\newcommand{\bluebegin}{\begingroup\color{blue}}
\newcommand{\blueend}{\endgroup}

\fi

\allowdisplaybreaks[4]

\newcommand{\Vladimir}{Vladimir}
\newcommand{\DOT}{.}

\ifnotLATIN
\input{OT2enc.def}

\usepackage{CJK}
\fi

\newcommand{\st}{\mathrel{\!|\!}}
\newcommand{\D}{\,\mathrm{d}}
\newcommand{\dd}{\mathrm{d}}

\newcommand{\K}{\mathcal{K}}		
\newcommand{\kkk}{\mathbf{k}}		
\newcommand{\ccc}{\mathbf{c}}		
\newcommand{\III}{\mathbb{I}}
\newcommand{\FFF}{\mathcal{F}}		
\newcommand{\GGG}{\mathcal{G}}		
\newcommand{\HHH}{\mathcal{H}}		
\newcommand{\PPP}{\mathcal{P}}		
\newcommand{\SSS}{\mathcal{S}}		

\newcommand{\bbbp}{\mathbb{P}}		
\newcommand{\Prob}{\mathop{\bbbp}\nolimits}

\newcommand{\sign}{\mathop{\mathrm{sign}}\nolimits}
\newcommand{\risk}{\mathop{\mathrm{risk}}\nolimits}
\newcommand{\Int}{\mathop{\mathrm{Int}}\nolimits}

\newcommand{\bbbr}{\mathbb{R}}		

\newtheorem{lemma}{Lemma}

\newtheorem{corollary}{Corollary}
\newtheorem{remark}{Remark}
\newtheorem{theorem}{Theorem}
\newenvironment{proof}
  {\trivlist\item[\hskip\labelsep\textbf{Proof}]}
  {\endtrivlist}

\newenvironment{Proof}[1]
  {\trivlist\item[\hskip\labelsep\textit{Proof #1:}]}
  {\endtrivlist}
\newcommand{\boxforqed}{\rule{.3em}{1.5ex}}
\newcommand{\qedtext}{\unskip\nobreak\hfil
  \penalty50\hskip1em\null\nobreak\hfil\boxforqed
  \parfillskip=0pt\finalhyphendemerits=0\endgraf}
\newcommand{\qedmath}{\tag*{\boxforqed}}

\ifJOURNAL
\title{Predictions as statements and decisions}
\author{Vladimir Vovk\\[5mm]
 Computer Learning Research Centre\\
  Department of Computer Science\\
  Royal Holloway, University of London,
  Egham, Surrey TW20 0EX, UK\\
  \texttt{vovk@cs.rhul.ac.uk}}
\fi

\ifarXiv
\title{Predictions as statements and decisions\\(draft: comments welcome)}
\author{Vladimir Vovk\\
\texttt{vovk{\rm@}cs.rhul.ac.uk}\\
\texttt{http://vovk.net}}
\fi

\ifWP
\title{Predictions as statements and decisions}
\author{Vladimir Vovk}

\fi

\ifFULL
\title{Predictions as statements and decisions}
\author{Vladimir Vovk\\
\texttt{vovk{\rm@}cs.rhul.ac.uk}\\
\texttt{http://vovk.net}}
\fi

\begin{document}
\maketitle
\begin{abstract}
  Prediction is a complex notion,
  and different predictors (such as people, computer programs, and probabilistic theories)
  can pursue very different goals.
  In this paper I will review some popular kinds of prediction
  and argue that the theory of competitive on-line learning
  can benefit from the kinds of prediction that are now foreign to it.

  The standard goal for predictor in learning theory
  is to incur a small loss for a given loss function
  measuring the discrepancy between the predictions and the actual outcomes.
  Competitive on-line learning concentrates on a ``relative'' version of this goal:
  the predictor is to perform almost as well as the best strategies
  in a given benchmark class of prediction strategies.
  Such predictions can be interpreted as decisions made by a ``small'' decision maker
  (i.e., one whose decisions do not affect the future outcomes).

  Predictions, or \emph{probability forecasts},
  considered in the foundations of probability
  are statements rather than decisions;
  the loss function is replaced by a procedure for testing the forecasts.
  The two main approaches to the foundations of probability
  are measure-theoretic
  (as formulated by Kolmogorov)
  and game-theoretic
  (as developed by von Mises and Ville);
  the former is now dominant in mathematical probability theory,
  but the latter appears to be better adapted for uses
  in learning theory
  discussed in this paper.

  An important achievement of Kolmogorov's school of the foundations of probability
  was construction of a universal testing procedure
  and realization (Levin, 1976) that there exists a forecasting strategy
  that produces ideal forecasts.
  Levin's ideal forecasting strategy, however, is not computable.
  Its more practical versions can be obtained
  from the results of game-theoretic probability theory.
  For a wide class of forecasting protocols,
  it can be shown that
  for any computable game-theoretic law of probability
  there exists a computable forecasting strategy that produces ideal forecasts,
  as far as this law of probability is concerned.
  Choosing suitable laws of probability
  we can ensure that the forecasts agree with reality
  in requisite ways.


  Probability forecasts that are known to agree with reality
  can be used for making good decisions:
  the most straightforward procedure
  is to select decisions that are optimal under the forecasts
  (the principle of minimum expected loss).
  This gives, \emph{inter alia}, a powerful tool for competitive on-line learning;
  I will describe its use for designing prediction algorithms
  that satisfy the property of universal consistency
  and its more practical versions.

  In conclusion of the paper I will discuss some limitations of
  competitive on-line learning
  and possible directions of further research.


  \ifarXiv
     \tableofcontents
  \fi
  \ifWP
     \newpage\thispagestyle{empty}
  \fi
\end{abstract}

\section{Introduction}
\label{sec:introduction}

This paper is based on my invited talk at the 19th Annual Conference on Learning Theory
(Pittsburgh, PA, June 24, 2006).
In recent years COLT invited talks have tended to aim at establishing connections
between the traditional concerns of the learning community
and the work done by other communities
(such as game theory, statistics, information theory, and optimization).
Following this tradition,
I will argue that some ideas from the foundations of probability
can be fruitfully applied in competitive on-line learning.

In this paper I will use the following informal taxonomy of predictions
(reminiscent of Shafer's \cite{shafer:1990}, Figure 2,
taxonomy of probabilities):
\begin{description}
\item[D-predictions]
  are mere Decisions.
  They can never be true or false but can be good or bad.
  Their quality is typically evaluated with a loss function.
\item[S-predictions]
  are Statements about reality.
  They can be tested and, if found inadequate, rejected as false.
\item[F-predictions]
  (or Frequentist predictions)
  are intermediate between D-pre\-dic\-tions and S-predictions.
  They are successful if they match the fre\-quen\-cies of various observed events.
\end{description}
Traditionally,
learning theory in general and competitive on-line learning in particular
consider D-predictions.
I will start, in Section \ref{sec:D-asymptotic},
from a simple asymptotic result about D-predictions:
there exists a universally consistent on-line prediction algorithm
(randomized if the loss function is not required to be convex in the prediction).
Section \ref{sec:S} is devoted to S-prediction
and Section \ref{sec:F} to F-prediction.
We will see that S-prediction is more fundamental than,
and can serve as a tool for, F-prediction.
Section \ref{sec:D-idea} explains how F-prediction
(and so, indirectly, S-prediction)
is relevant for D-prediction.
In Section \ref{sec:D-bounds} I will prove the result
of Section \ref{sec:D-asymptotic} about universal consistency,
as well as its non-asymptotic version.

\section{Universal consistency}
\label{sec:D-asymptotic}

In all prediction protocols in this paper
every player can see the other players' moves made so far
(they are \emph{perfect-information} protocols).
The most basic one is:

\bigskip

\noindent
\textsc{Prediction protocol}\nopagebreak
\begin{tabbing}
  \qquad\=\qquad\=\qquad\kill
  FOR $n=1,2,\dots$:\\
  \> Reality announces $x_n\in\mathbf{X}$.\\
  \> Predictor announces $\gamma_n\in\Gamma$.\\
  \> Reality announces $y_n\in\mathbf{Y}$.\\
  END FOR.
\end{tabbing}

\noindent
At the beginning of each round $n$ Predictor is given some data $x_n$
relevant to predicting the following observation $y_n$;
$x_n$ may contain information about $n$
and the previous observations $y_{n-1},y_{n-2},\ldots$\,.
The data is taken from the \emph{data space} $\mathbf{X}$
and the observations from the \emph{observation space} $\mathbf{Y}$.
The predictions $\gamma_n$ are taken from the \emph{prediction space} $\Gamma$,
and a prediction's quality in light of the actual observation
is measured by a \emph{loss function}
$\lambda:\mathbf{X}\times\Gamma\times\mathbf{Y}\to\bbbr$.
This is how we formalize D-predictions.
The prediction protocol will sometimes be referred to as the ``prediction game''
(in general, ``protocol'' and ``game'' will be used as synonyms,
with a tendency to use ``protocol'' when the players' goals
are not clearly stated;
for example,
a prediction game is a prediction protocol
complemented by a loss function).

We will always assume that the data space $\mathbf{X}$,
the prediction space $\Gamma$,
and the observation space $\mathbf{Y}$
are non-empty topological spaces
and that the loss function $\lambda$ is continuous.
Moreover,
we are mainly interested in the case
where $\mathbf{X}$, $\Gamma$, and $\mathbf{Y}$ are locally compact metric spaces,
the prime examples being Euclidean spaces and their open and closed subsets.
Traditionally only loss functions $\lambda(x,\gamma,y)=\lambda(\gamma,y)$
that do not depend on $x$ are considered in learning theory,
and this case appears to be most useful and interesting.
The reader might prefer to concentrate on this case.

Predictor's total loss over the first $N$ rounds is
$\sum_{n=1}^N \lambda(x_n,\gamma_n,y_n)$.
As usual in competitive on-line prediction
(see \cite{cesabianchi/lugosi:2006} for a recent book-length review of the field),
Predictor competes with a wide range of \emph{prediction rules}
$D:\mathbf{X}\to\Gamma$.
The total loss of such a prediction rule is
$\sum_{n=1}^N \lambda(x_n,D(x_n),y_n)$,
and so Predictor's goal is to achieve
\begin{equation}\label{eq:typical}
  \sum_{n=1}^N
  \lambda(x_n,\gamma_n,y_n)
  \lessapprox
  \sum_{n=1}^N
  \lambda(x_n,D(x_n),y_n)
\end{equation}
for all $N=1,2,\ldots$ and as many prediction rules $D$ as possible.

Predictor's strategies in the prediction protocol will be called
\emph{on-line prediction algorithms} (or \emph{strategies}).

\begin{remark}\label{rem:Cover}
  {\em Some common prediction games are not about prediction at all,
  as this word is usually understood.
  For example, in Cover's game of sequential investment
  (\cite{cesabianchi/lugosi:2006}, Chapter 10) with $K$ stocks,
  \begin{multline*}
    \mathbf{Y}
    :=
    [0,\infty)^K,
    \quad
    \Gamma
    :=
    \bigl\{
      (g_1,\ldots,g_K)\in[0,\infty)^K
      \st
      g_1+\cdots+g_K=1
    \bigr\},\\
    \lambda\bigl((g_1,\ldots,g_K),(y_1,\ldots,y_K)\bigr)
    :=
    -\ln
    \sum_{k=1}^K
    g_k y_k.
  \end{multline*}
  (there is no $\mathbf{X}$;
  or, more formally, $\mathbf{X}$ consists of one element which is omitted from our notation).
  The observation $y$ is interpreted
  as the ratios of the closing to opening price of the $K$ stocks
  and the ``prediction'' $\gamma$ is the proportions of the investor's capital
  invested in different stocks at the beginning of the round.
  The loss function is the minus logarithmic increase in the investor's capital.
  In this example $\gamma$ can hardly be called a prediction:
  in fact it is a decision made by a small decision maker,
  i.e., decision maker whose actions do not affect Reality's future behavior
  (see Section \ref{sec:conclusion}
  for a further discussion of this aspect of competitive on-line prediction).
  For other games of this kind, see \cite{vovk/watkins:1998}.}
\end{remark}

\subsection*{Universal consistency for deterministic prediction algorithms}

Let us say that a set in a topological space is \emph{precompact}
if its closure is compact.
In Euclidean spaces,
precompactness means boundedness.
An on-line prediction algorithm is \emph{universally consistent}
for a loss function $\lambda$
if its predictions $\gamma_n$ always satisfy
\begin{multline}\label{eq:universal-consistency}
  \bigl(
    \{x_1,x_2,\ldots\}
    \text{ and }
    \{y_1,y_2,\ldots\}
    \text{ are precompact}
  \bigr)\\
  \Longrightarrow
  \limsup_{N\to\infty}
  \left(
    \frac1N
    \sum_{n=1}^N
    \lambda(x_n,\gamma_n,y_n)
    -
    \frac1N
    \sum_{n=1}^N
    \lambda(x_n,D(x_n),y_n)
  \right)
  \le
  0
\end{multline}
for any continuous prediction rule $D:\mathbf{X}\to\Gamma$.
The intuition behind the antecedent of (\ref{eq:universal-consistency}),
in the Euclidean case,
is that the prediction algorithm
knows that $\left\|x_n\right\|$ and $\left\|y_n\right\|$ are bounded
but does not know an upper bound in advance.
Of course,
universal consistency is only a minimal requirement for successful prediction;
we will also be interested in bounds on the predictive performance
of our algorithms.

Let us say that the loss function $\lambda$ is \emph{compact-type}
if for each pair of compact sets $A\subseteq\mathbf{X}$ and $B\subseteq\mathbf{Y}$
and each constant $M$
there exists a compact set $C\subseteq\Gamma$ such that
\begin{equation*}
  \forall x\in A,\gamma\notin C,y\in B:
  \quad
  \lambda(x,\gamma,y)
  >
  M.
\end{equation*}
More intuitively,
we require that $\lambda(x,\gamma,y)\to\infty$ as $\gamma\to\infty$
uniformly in $(x,y)$ ranging over a compact set.
\begin{theorem}\label{thm:decision-asymptotic}
  Suppose $\mathbf{X}$ and $\mathbf{Y}$ are locally compact metric spaces,
  $\Gamma$ is a convex subset of a Fr\'echet space,
  and the loss function $\lambda(x,\gamma,y)$ is continuous,
  compact-type, and convex in the variable $\gamma\in\Gamma$.
  There exists a universally consistent on-line prediction algorithm.
\end{theorem}
To have a specific example in mind,
the reader might check that $\mathbf{X}=\bbbr^{K}$, $\Gamma=\mathbf{Y}=\bbbr^{L}$,
and $\lambda(x,\gamma,y):=\left\|y-\gamma\right\|$
satisfy the conditions of the theorem.

\subsection*{Universal consistency for randomized prediction algorithms}

When the loss function $\lambda(x,\gamma,y)$ is not convex in $\gamma$,
two difficulties appear:
\begin{itemize}
\item
  the conclusion of Theorem \ref{thm:decision-asymptotic} becomes false
  if the convexity requirement is removed
  (\cite{kalnishkan/vyugin:2005}, Theorem 2);
\item
  in some cases the notion of a continuous prediction rule becomes vacuous:
  e.g., there are no non-constant continuous prediction rules
  when $\Gamma=\{0,1\}$ and $\mathbf{X}$ is connected.
\end{itemize}
To overcome these difficulties,
we consider randomized prediction rules and randomized on-line prediction algorithms
(with independent randomizations).
It will follow from the proof of Theorem \ref{thm:decision-asymptotic}
that one can still guarantee that (\ref{eq:universal-consistency}) holds,
although with probability one;
on the other hand,
there will be a vast supply of continuous prediction rules.

\begin{remark}
  \emph{In fact,
  the second difficulty is more apparent than real:
  for example, in the binary case ($\mathbf{Y}=\{0,1\}$)
  with the loss function $\lambda(\gamma,y)$ independent of $x$,
  there are many non-trivial continuous prediction rules
  in the canonical form of the prediction game \cite{vovk:1990}
  with the prediction set redefined as the boundary of the set of superpredictions
  \cite{kalnishkan/vyugin:2005}.}
\end{remark}

A \emph{randomized prediction rule} is a function $D:\mathbf{X}\to\PPP(\Gamma)$
mapping the data space into the probability measures on the prediction space;
$\PPP(\Gamma)$ is always equipped with the topology of weak convergence
\cite{billingsley:1968}.
A \emph{randomized on-line prediction algorithm} is an on-line prediction algorithm
in the extended prediction game with the prediction space $\PPP(\Gamma)$.
Let us say that a randomized on-line prediction algorithm is \emph{universally consistent}
if,
for any continuous randomized prediction rule $D:\mathbf{X}\to\PPP(\Gamma)$,
\begin{multline}\label{eq:universal-consistency-randomized}
  \bigl(
    \{x_1,x_2,\ldots\}
    \text{ and }
    \{y_1,y_2,\ldots\}
    \text{ are precompact}
  \bigr)\\
  \Longrightarrow
  \left(
    \limsup_{N\to\infty}
    \left(
      \frac1N
      \sum_{n=1}^N
      \lambda(x_n,g_n,y_n)
      -
      \frac1N
      \sum_{n=1}^N
      \lambda(x_n,d_n,y_n)
    \right)
    \le
    0
    \enspace
    \textrm{a.s.}
  \right)
\end{multline}
where $g_1,g_2,\ldots,d_1,d_2,\ldots$ are independent random variables
with $g_n$ distributed as $\gamma_n$ and $d_n$ distributed as $D(x_n)$,
$n=1,2,\ldots$\,.
Intuitively,
the ``a.s.''\ in (\ref{eq:universal-consistency-randomized})
refers to the algorithm's and prediction rule's internal randomization.
\begin{theorem}\label{cor:decision-asymptotic}
  Let $\mathbf{X}$ and $\mathbf{Y}$ be locally compact metric spaces,
  $\Gamma$ be a metric space,
  and $\lambda$ be a continuous and compact-type loss function.
  There exists a universally consistent randomized on-line prediction algorithm.
\end{theorem}

Let $\mathbf{X}$ be a metric space.
For any discrete (e.g., finite) subset $\{x_1,x_2,\ldots\}$ of $\mathbf{X}$
and any sequence $\gamma_n\in\PPP(\Gamma)$ of probability measures on $\Gamma$
there exists a continuous randomized prediction rule $D$
such that $D(x_n)=\gamma_n$ for all $n$
(indeed, it suffices to set $D(x):=\sum_n\phi_n(x)\gamma_n$,
where $\phi_n:\mathbf{X}\to[0,1]$, $n=1,2,\ldots$,
are continuous functions with disjoint supports
such that $\phi_n(x_n)=1$ for all $n$).
Therefore, there is no shortage of randomized prediction rules.

\subsection*{Continuity, compactness, and the statistical notion of universal consistency}

In the statistical setting,
where $(x_n,y_n)$ are assumed to be generated independently
from the same probability measure,
the definition of universal consistency was given by Stone \cite{stone:1977} in 1977.
One difference of Stone's definition from ours
is the lack of the requirement that $D$ should be continuous
in his definition.

If the requirement of continuity of $D$ is dropped
from our definition,
universal consistency becomes impossible to achieve:
Reality can easily choose $x_n\to c$, where $c$ is a point of discontinuity of $D$,
and $y_n$ in such a way that Predictor's loss will inevitably be much larger than $D$'s.
To be more specific,
suppose $\mathbf{X}=\Gamma=\mathbf{Y}=[-1,1]$
and $\lambda(x,\gamma,y)=\left|y-\gamma\right|$
(more generally, the loss is zero when $y=\gamma$ and positive when $y\ne\gamma$).
No matter how Predictor chooses his predictions $\gamma_n$,
Reality can choose
\begin{equation*}
  x_n
  :=
  \sum_{i=1}^{n-1}
  \frac{\sign\gamma_i}{3^i},
  \quad
  y_n
  :=
  -\sign\gamma_n,
\end{equation*}
where the function $\sign$ is defined as
\begin{equation*}
  \sign\gamma
  :=
  \begin{cases}
    1 & \text{if $\gamma\ge0$}\\
    -1 & \text{otherwise},
  \end{cases}
\end{equation*}
and thus foil (\ref{eq:universal-consistency}) for the prediction rule
\begin{equation*}
  D(x)
  :=
  \begin{cases}
    -1 & \text{if $x<\sum_{i=1}^{\infty} (\sign\gamma_i)/3^i$}\\
    1 & \text{otherwise}.
  \end{cases}
\end{equation*}
(Indeed, these definitions imply $D(x_n)=-\sign\gamma_n=y_n$ for all $n$.)

A positive argument in favor of the requirement of continuity of $D$
is that it is natural for Predictor to compete only
with computable prediction rules,
and continuity is often regarded as a necessary condition for computability
(Brouwer's ``continuity principle'').

Another difference of Stone's definition
is that compactness does not play any special role in it
(cf.\ the antecedent of (\ref{eq:universal-consistency})).
It is easy to see that the condition that $\{x_1,x_2,\ldots\}$ and $\{y_1,y_2,\ldots\}$
are precompact is essential in our framework.
Indeed, let us suppose, e.g., that $\{x_1,x_2,\ldots\}$ is allowed not to be precompact,
continuing to assume that $\mathbf{X}$ is a metric space
and also assuming that $\mathbf{Y}$ is a convex subset of a topological vector space.
Reality can then choose $x_n$, $n=1,2,\ldots$, as a discrete set in $\mathbf{X}$
(\cite{engelking:1989}, 4.1.17).
Let $\phi_n:\mathbf{X}\to[0,1]$, $n=1,2,\ldots$,
be continuous functions with disjoint supports
such that $\phi_n(x_n)=1$ for all $n$.
For any sequence of observations $y_1,y_2,\ldots$,
the function $D(x):=\sum_n\phi_n(x)y_n$ is a continuous prediction rule
such that $D(x_n)=y_n$ for all $n$.
Under such circumstances
it is impossible to compete with all continuous prediction rules
unless the loss function satisfies some very special properties.

As compared to competitive on-line prediction,
the statistical setting is rather restrictive.
Compactness and continuity may be said to be satisfied automatically:
under mild conditions,
every measurable prediction rule can be arbitrarily well approximated
by a continuous one
(according to Luzin's theorem,
\cite{dudley:2002}, 7.5.2,
combined with the Tietze--Uryson theorem, \cite{engelking:1989}, 2.1.8),
and every probability measure is almost concentrated on a compact set
(according to Ulam's theorem,
\cite{dudley:2002}, 7.1.4).

\section{Defensive forecasting}
\label{sec:S}

In this and next sections we will discuss S-prediction and F-prediction,
which will prepare way for proving
Theorems \ref{thm:decision-asymptotic} and \ref{cor:decision-asymptotic}.
\begin{remark}
  {\em In this paper,
  S-predictions and F-predictions will always be probability measures,
  whereas typical D-predictions are not measures.
  This difference is, however, accidental:
  e.g., in the problem of on-line regression
  (as in \cite{vovk:leading}, Section 5)
  different kinds of predictions are objects of the same nature.}
\end{remark}

\subsection*{Testing predictions in measure-theoretic probability
  and neutral measures}

S-predictions are empirical statements about the future;
they may turn out true or false
as the time passes.
For such statements to be non-vacuous,
we need to have a clear idea of when they become falsified
by future observations
\cite{popper:1934}.
In principle,
the issuer of S-predictions should agree in advance
to a protocol of testing his predictions.
It can be said that such a protocol provides an empirical meaning to the predictions.

Testing is, of course, a well-developed area of statistics
(see, e.g., \cite{cox/hinkley:1974}, Chapter 3).
A typical problem is:
given a probability measure (the ``null hypothesis'') $P$ on a set $\Omega$,
which observations $\omega\in\Omega$ falsify $P$?
In the context of this paper,
$P$ is an S-prediction,
or, as we will often say, a probability forecast for $\omega\in\Omega$.
Developing Kolmogorov's ideas
(see, e.g., \cite{kolmogorov:1965}, Section 4, \cite{kolmogorov:1968}, and \cite{kolmogorov:1983}),
Martin-L\"of (1966, \cite{martin-lof:1966})
defines a (in some sense, ``the'') universal statistical test
for a computable $P$.
Levin (1976, \cite{\Levin}) modifies Martin-L\"of's definition of statistical test
(which was, in essence, the standard statistical definition)
and extends it to noncomputable $P$;
Levin's 1976 definition is ``uniform'', in an important sense.

Levin's test is a function $t:\Omega\times\PPP(\Omega)\to[0,\infty]$,
where $\PPP(\Omega)$ is the set of all Borel probability measures on $\Omega$,
assumed to be a topological space.
Levin \cite{\Levin} considers the case $\Omega=\{0,1\}^{\infty}$
but notes that his argument works for any other ``good'' compact space
with a countable base.
We will assume that $\Omega$ is a metric compact
(which is equivalent to Levin's assumption that $\Omega$ is a compact space
with a countable base, \cite{engelking:1989}, 4.2.8),
endowing $\PPP(\Omega)$ with the topology of weak convergence
(see below for references).
Let us say that a function
$t:\Omega\times\PPP(\Omega)\to[0,\infty]$
is a \emph{test of randomness} if it is lower semicontinuous
and, for all $P\in\PPP(\Omega)$,
\begin{equation*}
  \int_{\Omega}
  t(\omega,P)
  P(\dd\omega)
  \le
  1.
\end{equation*}
The intuition behind this definition is that
if we first choose a test $t$,
then observe $\omega$,
and then find that $t(\omega,P)$ is very large for the observed $\omega$,
we are entitled to reject the hypothesis that $\omega$ was generated from $P$
(notice that the $P$-probability that $t(\omega,P)\ge C$ cannot exceed $1/C$,
for any $C>0$).

The following fundamental result is due to Levin
(\cite{\Levin}, footnote ${}^{(1)}$),
although our proof is slightly different
(for details of Levin's proof, see \cite{gacs:2005}, Section 5).
\begin{lemma}[Levin]\label{lem:levin}
  Let $\Omega$ be a metric compact.
  For any test of randomness $t$ there exists a probability measure $P$
  such that
  \begin{equation}\label{eq:neutral}
  \forall \omega\in\Omega:
  \quad
  t(\omega,P)\le1.
  \end{equation}
\end{lemma}
Before proving this result,
let us recall some useful facts about the probability measures
on the metric compact $\Omega$.
The Banach space of all continuous functions on $\Omega$
with the usual pointwise addition and scalar action
and the sup norm will be denoted $C(\Omega)$.
By one of the Riesz representation theorems
(\cite{dudley:2002}, 7.4.1; see also 7.1.1),
the mapping $\mu\mapsto I_{\mu}$,
where
$
  I_{\mu}(f):=\int_{\Omega}f\D\mu
$,
is a linear isometry
between the set of all finite Borel measures $\mu$ on $\Omega$
with the total variation norm
and the dual space $C'(\Omega)$ to $C(\Omega)$
with the standard dual norm
(\cite{rudin:1991}, Chapter 4).
We will identify the finite Borel measures $\mu$ on $\Omega$
with the corresponding $I_{\mu}\in C'(\Omega)$.
This makes $\PPP(\Omega)$ a convex closed subset of $C'(\Omega)$.

We will be interested, however,
in a different topology on $C'(\Omega)$,
the weakest topology for which all evaluation functionals
$\mu\in C'(\Omega)\mapsto\mu(f)$, $f\in C(\Omega)$,
are continuous.
This topology is known as the \emph{weak${}^*$ topology}
(\cite{rudin:1991}, 3.14),
and the topology inherited by $\PPP(\Omega)$
is known as the \emph{topology of weak convergence}
(\cite{billingsley:1968}, Appendix III).
The point mass $\delta_{\omega}$, $\omega\in\Omega$,
is defined to be the probability measure concentrated at $\omega$,
$\delta_{\omega}(\{\omega\})=1$.
The simple example of a sequence of point masses $\delta_{\omega_n}$
such that $\omega_n\to\omega$ as $n\to\infty$ and $\omega_n\ne\omega$ for all $n$
shows that the topology of weak convergence is different from the dual norm topology:
$\delta_{\omega_n}\to\delta_{\omega}$ holds in one but does not hold in the other.

It is not difficult to check that $\PPP(\Omega)$ remains a closed subset of $C'(\Omega)$
in the weak${}^*$ topology
(\cite{bourbaki:integration}, III.2.7, Proposition 7).
By the Banach--Alaoglu theorem
(\cite{rudin:1991}, 3.15)
$\PPP(\Omega)$ is compact in the topology of weak convergence
(this is a special case of Prokhorov's theorem,
\cite{billingsley:1968}, Appendix III, Theorem 6).
In the rest of this paper,
$\PPP(\Omega)$
(and all other spaces of probability measures)
are always equipped with the topology of weak convergence.

Since $\Omega$ is a metric compact,
$\PPP(\Omega)$ is also metrizable
(by the well-known Prokhorov metric:
\cite{billingsley:1968}, Appendix III, Theorem 6).

\begin{Proof}{of Lemma \ref{lem:levin}}
  If $t$ takes value $\infty$,
  redefine it as $t:=\min(t,2)$.
  For all $P,Q\in\PPP(\Omega)$
  set
  \begin{equation*}
    \phi(Q,P)
    :=
    \int_{\Omega}
      t(\omega,P)
    Q(\dd\omega).
  \end{equation*}
  The function $\phi(Q,P)$ is linear in its first argument, $Q$,
  and lower semicontinuous (see Lemma~\ref{lem:semicontinuity} below)
  in its second argument, $P$.
  Ky Fan's minimax theorem
  (see, e.g., \cite{agarwal/etal:2001}, Theorem 11.4;
  remember that $\PPP(\Omega)$ is a compact convex subset of $C'(\Omega)$
  equipped with the weak${}^*$ topology)
  shows that there exists $P^*\in\PPP(\Omega)$ such that
  \begin{equation*}
    \forall Q\in\PPP(\Omega):
    \quad
    \phi(Q,P^*)
    \le
    \sup_{P\in\PPP(\Omega)}
    \phi(P,P).
  \end{equation*}
  Therefore,
  \begin{equation*}
    \forall Q\in\PPP(\Omega):
    \quad
    \int_{\Omega}
      t(\omega,P^*)
    Q(\dd\omega)
    \le
    1,
  \end{equation*}
  and we can see that $t(\omega,P^*)$ never exceeds $1$.
  \qedtext
\end{Proof}
This proof used the following topological lemma.
\begin{lemma}\label{lem:semicontinuity}
  Suppose $F:X\times Y\to\bbbr$ is a non-negative lower semicontinuous function
  defined on the product of two metric compacts, $X$ and $Y$.
  If $Q$ is a probability measure on $Y$,
  the function $x\in X\mapsto\int_Y F(x,y)Q(\dd y)$ is also lower semicontinuous.
\end{lemma}
\begin{proof}
  The product $X\times Y$ is also a metric compact
  (\cite{engelking:1989}, 3.2.4 and 4.2.2).
  According to Hahn's theorem
  (\cite{engelking:1989}, Problem 1.7.15(c)),
  there exists a non-decreasing sequence of (non-negative) continuous functions $F_n(x,y)$
  such that $F_n(x,y)\to F(x,y)$ as $n\to\infty$ for all $(x,y)\in X\times Y$.
  Since each $F_n$ is uniformly continuous
  (\cite{engelking:1989}, 4.3.32),
  the functions $\int_Y F_n(x,y)Q(\dd y)$ are continuous,
  and by the monotone convergence theorem
  (\cite{dudley:2002}, 4.3.2)
  they converge to $\int_Y F(x,y)Q(\dd y)$.
  Therefore, again by Hahn's theorem,
  $\int_Y F(x,y)Q(\dd y)$ is lower semicontinuous.
  \qedtext
\end{proof}

Lemma \ref{lem:levin} says that for any test of randomness $t$
there is a probability forecast $P$ such that $t$ never detects
any disagreement between $P$ and the outcome $\omega$,
whatever $\omega$ might be.

Gacs (\cite{gacs:2005}, Section 3) defines a uniform test of randomness
as a test of randomness that is lower semicomputable
(lower semicomputability is an ``effective'' version of the requirement of lower semicontinuity;
this requirement is very natural in the context of randomness:
cf.\ \cite{vovk/vyugin:1993}, Section 3.1).
He proves (\cite{gacs:2005}, Theorem 1) that there exists
a \emph{universal} (i.e., largest to within a constant factor)
uniform test of randomness.
If $t(\omega,P)<\infty$ for a fixed universal test $t$,
$\omega$ is said to be \emph{random} with respect to $P$.
Applied to the universal test,
Lemma \ref{lem:levin} says that there exists a ``neutral'' probability measure $P$,
such that every $\omega$ is random with respect to $P$.

Gacs (\cite{gacs:2005}, Theorem 7) shows that under his definition
there are no neutral measures that are computable even in the weak sense
of upper or lower semicomputability even for $\Omega$ the compactified set of natural numbers.
Levin's original definition of a uniform test of randomness
involved some extra conditions,
which somewhat mitigate (but not solve completely) the problem of non-computability.

\subsection*{Testing predictions in game-theoretic probability}

There is an obvious mismatch
between the dynamic prediction protocol of Section \ref{sec:D-asymptotic}
and the one-step probability forecasting setting of the previous subsection.
If we still want to fit the former into the latter,
perhaps we will have to take the infinite sequence of data and observations,
$x_1,y_1,x_2,y_2,\ldots$,
as $\omega$,
and so take $\Omega:=(\mathbf{X}\times\mathbf{Y})^{\infty}$.
To find a probability measure
satisfying a useful property,
such as (\ref{eq:neutral}) for an interesting $t$,
might be computationally expensive.
Besides,
this would force us to assume that the $x_n$s are also generated from $P$,
and it would be preferable to keep them free of any probabilities
(we cannot assume that $x_n$ are given constants
since they, e.g., may depend on the previous observations).

A more convenient framework is provided by the game-theoretic foundations of probability.
This framework was first thoroughly explored by von Mises \cite{mises:1919,mises:1928}
(see \cite{shafer/vovk:2001}, Chapter 2, for von Mises's precursors),
and a serious shortcoming of von Mises's theory was corrected by Ville \cite{ville:1939}.
After Ville,
game-theoretic probability was dormant
before being taken up by Kolmogorov \cite{kolmogorov:1968,kolmogorov:1983}.
The independence of game-theoretic probability
from the standard measure-theoretic probability \cite{kolmogorov:1933}
was emphasized by Dawid
(cf.\ his prequential principle in \cite{dawid:1984,dawid:1986});
see \cite{shafer/vovk:2001} for a review.

There is a special player in the game-theoretic protocols
who is responsible for testing the forecasts;
following \cite{shafer/vovk:2001},
this player will be called Skeptic.
This is the protocol that we will be using in this paper:

\bigskip

\noindent
\textsc{Testing protocol}\nopagebreak
\begin{tabbing}
  \qquad\=\qquad\=\qquad\kill
  FOR $n=1,2,\dots$:\\
  \> Reality announces $x_n\in\mathbf{X}$.\\
  \> Forecaster announces $P_n\in\PPP(\mathbf{Y})$.\\
  \> Skeptic announces $f_n:\mathbf{Y}\to\bbbr$ such that $\int_{\mathbf{Y}}f_n\D P_n\le0$.\\
  \> Reality announces $y_n\in\mathbf{Y}$.\\
  \> $\K_n := \K_{n-1} + f_n(y_n)$.\\
  END FOR.
\end{tabbing}

\noindent
Skeptic's move $f_n$ can be interpreted as taking a long position in a security
that pays $f_n(y_n)$ after $y_n$ becomes known;
according to Forecaster's beliefs encapsulated in $P_n$,
Skeptic does not have to pay anything for this.
We write $\int_{\mathbf{Y}}f_n\D P_n\le0$ to mean
that $\int_{\mathbf{Y}}f_n\D P_n$ exists and is non-positive.
Skeptic starts from some initial capital $\K_0$,
which is not specified in the protocol;
the evolution of $\K_n$, however, is described.

A game-theoretic procedure of testing Forecaster's performance
is a strategy for Skeptic in the testing protocol.
If Skeptic starts from $\K_0:=1$,
plays so that he never risks bankruptcy
(we say that he risks bankruptcy
if his move $f_n$
makes it possible for Reality to choose $y_n$ making $\K_n$ negative),
and ends up with a very large value $\K_N$ of his capital,
we are entitled to reject the forecasts as false.
Informally,
the role of Skeptic is to detect disagreement between the forecasts
and the actual observations,
and the current size of his capital tells us how successful he is
at achieving this goal.

\subsection*{Defensive forecasting}

Levin's Lemma \ref{lem:levin}
can be applied to any testing procedure $t$
(test of randomness)
to produce forecasts that are ideal
as far as that testing procedure is concerned.
Such ideal forecasts will be called ``defensive forecasts'';
in this subsection we will be discussing
a similar procedure of defensive forecasting
in game-theoretic probability.

Let us now slightly change the testing protocol:
suppose that right after Reality's first move in each round
Skeptic announces his strategy for the rest of that round.

\bigskip

\noindent
\textsc{Defensive forecasting protocol}\nopagebreak
\begin{tabbing}
  \qquad\=\qquad\=\qquad\kill
  FOR $n=1,2,\dots$:\\
  \> Reality announces $x_n\in\mathbf{X}$.\\
  \> Skeptic announces a lower semicontinuous $F_n:\mathbf{Y}\times\PPP(\mathbf{Y})\to\bbbr$\\
  \>\> such that $\int_{\mathbf{Y}}F_n(y,P)P(\dd y)\le0$ for all $P\in\PPP(\mathbf{Y})$.\\
  \> Forecaster announces $P_n\in\PPP(\mathbf{Y})$.\\
  \> Reality announces $y_n\in\mathbf{Y}$.\\
  \> $\K_n := \K_{n-1} + F_n(y_n,P_n)$.\\
  END FOR.
\end{tabbing}

\noindent
This protocol will be used in the situation
where Skeptic has chosen in advance, and told Forecaster about, his testing strategy.
However, the game-theoretic analogue of Levin's lemma
holds even when Skeptic's strategy is disclosed in a piecemeal manner,
as in our protocol.

The following lemma can be proven in the same way as
(and is a simple corollary of)
Levin's Lemma \ref{lem:levin}.
Its version was first obtained by Akimichi Takemura in 2004 \cite{\Takemura}.
\begin{lemma}[Takemura]\label{lem:takemura}
  Let $\mathbf{Y}$ be a metric compact.
  In the defensive forecasting protocol,
  Forecaster can play in such a way that Skeptic's capital never increases,
  no matter how he and Reality play.
\end{lemma}
\begin{proof}
  For all $P,Q\in\PPP(\mathbf{Y})$
  set
  \begin{equation*}
    \phi(Q,P)
    :=
    \int_{\mathbf{Y}}
      F_n(y,P)
    Q(\dd y),
  \end{equation*}
  where $F_n$ is Skeptic's move in round $n$.
  The function $\phi(Q,P)$ is linear in $Q$ and lower semicontinuous in $P$
  (the latter also follows from Lemma \ref{lem:semicontinuity}
  if we notice that the assumption that $F$ is non-negative can be removed:
  every lower semicontinuous function on a compact set is bounded below).
  Ky Fan's minimax theorem
  shows that there exists $P^*$ such that
  \begin{equation*}
    \phi(Q,P^*)
    \le
    \sup_{P\in\PPP(\mathbf{Y})}
    \phi(P,P)
    \le
    0,
  \end{equation*}
  and we can see that $F_n(y,P^*)$ is always non-positive.
  Since the increment $\K_n-\K_{n-1}$ equals $F_n(y_n,P_n)$,
  it suffices to set $P_n:=P^*$.
  \qedtext
\end{proof}

\subsection*{Testing and laws of probability}

There are many interesting ways of testing probability forecasts.
In fact,
every law of probability provides a way of testing probability forecasts
(and vice versa,
any way of testing probability forecasts
can be regarded as a law of probability).
As a simple example,
consider the strong law of large numbers in the binary case
($\mathbf{Y}=\{0,1\}$):
\begin{equation}\label{eq:SLLN}
  \lim_{N\to\infty}
  \frac1N
  \sum_{n=1}^N
  (y_n-p_n)
  =
  0
\end{equation}
with probability one,
where $p_n:=P_n(\{1\})$ is the predicted probability that $y_n=1$.
If (\ref{eq:SLLN}) is violated,
we are justified in rejecting the forecasts $p_n$;
in this sense the strong law of large numbers can serve as a test.

In game-theoretic probability theory,
the binary strong law of large numbers is stated as follows:
Skeptic has a strategy that,
when started with $\K_0:=1$,
never risks bankruptcy and makes Skeptic infinitely rich when (\ref{eq:SLLN})
is violated.
We prove many such game-theoretic laws of probability
in \cite{shafer/vovk:2001};
all of them exhibit strategies (continuous or easily made continuous)
for Skeptic that make him rich
when some property of agreement
(such as, apart from various laws of large numbers,
the law of the iterated logarithm and the central limit theorem)
between the forecasts and the actual observations
is violated.
When Forecaster plays the strategy of defensive forecasting
against such a strategy for Skeptic,
the property of agreement is guaranteed to be satisfied,
no matter how Reality plays.

In the next section we will apply the procedure of defensive forecasting
to a law of large numbers found by Kolmogorov in 1929
(\cite{kolmogorov:1929LLN};
its simple game-theoretic version can be found in \cite{shafer/vovk:2001},
Lemma 6.1 and Proposition 6.1).

\section{Calibration and resolution}
\label{sec:F}

In this section we will see how the idea of defensive forecasting
can be used for producing F-predictions.
It is interesting that the pioneering work in this direction
by Foster and Vohra \cite{foster/vohra:1998}
was completely independent of Levin's idea.
The following is our basic probability forecasting protocol
(more basic than the protocols of the previous section).

\bigskip

\noindent
\textsc{Probability forecasting protocol}\nopagebreak
\begin{tabbing}
  \qquad\=\qquad\=\qquad\kill
  FOR $n=1,2,\dots$:\\
  \> Reality announces $x_n\in\mathbf{X}$.\\
  \> Forecaster announces $P_n\in\PPP(\mathbf{Y})$.\\
  \> Reality announces $y_n\in\mathbf{Y}$.\\
  END FOR.
\end{tabbing}

\noindent
Forecaster's prediction $P_n$ is a probability measure on $\mathbf{Y}$
that, intuitively, describes his beliefs about the likely values of $y_n$.
Forecaster's strategy in this protocol
will be called a \emph{probability forecasting strategy}
(or \emph{algorithm}).

\subsection*{Asymptotic theory of calibration and resolution}

The following is a simple asymptotic result
about the possibility to ensure ``calibration'' and ``resolution''.
\begin{theorem}\label{thm:forecasting-asymptotic}
  Suppose $\mathbf{X}$ and $\mathbf{Y}$ are locally compact metric spaces.
  There is a probability forecasting strategy that guarantees
  \begin{multline}\label{eq:calibration-cum-resolution}
    \bigl(
      \{x_1,x_2,\ldots\}
      \text{ and }
      \{y_1,y_2,\ldots\}
      \text{ are precompact}
    \bigr)\\
    \Longrightarrow
    \lim_{N\to\infty}
    \frac1N
    \sum_{n=1}^N
    \left(
      f
      \left(
        x_n,P_n,y_n
      \right)
      -
      \int_{\mathbf{Y}}
      f
      \left(
        x_n,P_n,y
      \right)
      P_n(\dd y)
    \right)
    =
    0
  \end{multline}
  for all continuous functions $f:\mathbf{X}\times\PPP(\mathbf{Y})\times\mathbf{Y}\to\bbbr$.
\end{theorem}

This theorem will be proven at the end of this section,
and in the rest of this subsection I will explain
the intuition behind (\ref{eq:calibration-cum-resolution}).
The discussion here is an extension of that in \cite{\GTPXIII}, Section 6.
Let us assume, for simplicity,
that $\mathbf{X}$ and $\mathbf{Y}$ are compact metric spaces;
as before,
$\delta_y$, where $y\in\mathbf{Y}$,
stands for the probability measure in $\PPP(\mathbf{Y})$
concentrated on $\{y\}$.

We start from the intuitive notion of calibration
(for further details, see \cite{dawid:1986} and \cite{foster/vohra:1998}).
The probability forecasts $P_n$, $n=1,\ldots,N$,
are said to be ``well calibrated''
(or ``unbiased in the small'', or ``reliable'', or ``valid'')
if, for any $P^*\in\PPP(\mathbf{Y})$,
\begin{equation}\label{eq:calibration-1}
  \frac
  {\sum_{n=1,\ldots,N:P_n\approx P^*} \delta_{y_n}}
  {\sum_{n=1,\ldots,N:P_n\approx P^*} 1}
  \approx
  P^*
\end{equation}
provided $\sum_{n=1,\ldots,N:P_n\approx P^*} 1$ is not too small.
The interpretation of (\ref{eq:calibration-1}) is that the forecasts
should be in agreement with the observed frequencies.
We can rewrite (\ref{eq:calibration-1}) as
\begin{equation*}
  \frac
  {\sum_{n=1,\ldots,N:P_n\approx P^*} (\delta_{y_n}-P_n)}
  {\sum_{n=1,\ldots,N:P_n\approx P^*} 1}
  \approx
  0.
\end{equation*}
Assuming that $P_n\approx P^*$ for a significant fraction of the $n=1,\ldots,N$,
we can further restate this as the requirement that
\begin{equation}\label{eq:calibration-2}
  \frac1N
  \sum_{n=1,\ldots,N:P_n\approx P^*}
  \left(
    g(y_n)
    -
    \int_{\mathbf{Y}} g(y) P_n(\dd y)
  \right)
  \approx
  0
\end{equation}
for a wide range of continuous functions $g$
(cf.\ the definition of the topology of weak convergence in the previous section).

The fact that good calibration is only a necessary condition
for good forecasting performance
can be seen from the following standard example
\cite{dawid:1986,foster/vohra:1998}:
if $\mathbf{Y}=\{0,1\}$ and
\begin{equation*}
  (y_1,y_2,y_3,y_4,\ldots)
  =
  (1,0,1,0,\ldots),
\end{equation*}
the forecasts $P_n(\{0\})=P_n(\{1\})=1/2$, $n=1,2,\ldots$,
are well calibrated but rather poor;
it would be better to forecast with
\begin{equation*}
  (P_1,P_2,P_3,P_4,\ldots)
  =
  (\delta_1,\delta_0,\delta_1,\delta_0,\ldots).
\end{equation*}
Assuming that each datum $x_n$ contains the information about the parity of $n$
(which can always be added to $x_n$),
we can see that the problem with the former forecasting strategy
is its lack of resolution:
it does not distinguish between the data with odd and even $n$.
In general, we would like each forecast $P_n$
to be as specific as possible to the current datum $x_n$;
the resolution of a probability forecasting algorithm
is the degree to which it achieves this goal
(taking it for granted that $x_n$ contains all relevant information).

Analogously to (\ref{eq:calibration-2}),
the forecasts $P_n$, $n=1,\ldots,N$, may be said to have good resolution
if, for any $x^*\in\mathbf{X}$,
\begin{equation}\label{eq:resolution}
  \frac1N
  \sum_{n=1,\ldots,N:x_n\approx x^*}
  \left(
    g(y_n)
    -
    \int_{\mathbf{Y}} g(y) P_n(\dd y)
  \right)
  \approx
  0
\end{equation}
for a wide range of continuous $g$.
We can also require that the forecasts $P_n$, $n=1,\ldots,N$,
should have good ``calibration-cum-resolution'':
for any $(x^*,P^*)\in\mathbf{X}\times\PPP(\mathbf{Y})$,
\begin{equation}\label{eq:combination}
  \frac1N
  \sum_{n=1,\ldots,N:(x_n,P_n)\approx(x^*,P^*)}
  \left(
    g(y_n)
    -
    \int_{\mathbf{Y}} g(y) P_n(\dd y)
  \right)
  \approx
  0
\end{equation}
for a wide range of continuous $g$.
Notice that even if forecasts have both good calibration and good resolution,
they can still have poor calibration-cum-resolution.

To make sense of the $\approx$ in, say, (\ref{eq:calibration-2}),
we can replace each ``crisp'' point $P^*\in\PPP(\mathbf{Y})$
by a ``fuzzy point'' $I_{P^*}:\PPP(\mathbf{Y})\to[0,1]$;
$I_{P^*}$ is required to be continuous,
and we might also want to have $I_{P^*}(P^*)=1$
and $I_{P^*}(P)=0$ for all $P$ outside a small neighborhood of $P^*$.
(The alternative of choosing $I_{P^*}:=\III_{A}$,
where $A$ is a small neighborhood of $P^*$ and $\III_A$ is its indicator function,
does not work because of Oakes's and Dawid's examples
\cite{oakes:1985,dawid:1985JASA};
$I_{P^*}$ can, however,
be arbitrarily close to $\III_A$.)
This transforms (\ref{eq:calibration-2}) into
\begin{equation*}
  \frac1N
  \sum_{n=1}^N
  I_{P^*}(P_n)
  \left(
    g(y_n)
    -
    \int_{\mathbf{Y}} g(y) P_n(\dd y)
  \right)
  \approx
  0,
\end{equation*}
which is equivalent to
\begin{equation}\label{eq:calibration-3}
  \frac1N
  \sum_{n=1}^N
  \left(
    f(P_n,y_n)
    -
    \int_{\mathbf{Y}} f(P_n,y) P_n(\dd y)
  \right)
  \approx
  0,
\end{equation}
where
$f(P,y):=I_{P^*}(P)g(y)$.
It is natural to require that (\ref{eq:calibration-3}) should hold
for a wide range of continuous functions $f(P,y)$,
not necessarily of the form $I_{P^*}(P)g(y)$.

In the same way we can transform (\ref{eq:resolution}) into
\begin{equation*}
  \frac1N
  \sum_{n=1}^N
  \left(
    f(x_n,y_n)
    -
    \int_{\mathbf{Y}} f(x_n,y) P_n(\dd y)
  \right)
  \approx
  0
\end{equation*}
and (\ref{eq:combination}) into
\begin{equation*}
  \frac1N
  \sum_{n=1}^N
  \left(
    f(x_n,P_n,y_n)
    -
    \int_{\mathbf{Y}} f(x_n,P_n,y) P_n(\dd y)
  \right)
  \approx
  0.
\end{equation*}
We can see that the consequent of (\ref{eq:calibration-cum-resolution})
can be interpreted as the forecasts having good calibration-cum-resolution;
the case where $f(x,P,y)$ depends only on $P$ and $y$
corresponds to good calibration,
and the case where $f(x,P,y)$ depends only on $x$ and $y$
corresponds to good resolution.

\subsection*{Calibration-cum-resolution bounds}

A more explicit result about calibration and resolution
is given in terms of ``reproducing kernel Hilbert spaces''.
Let $\FFF$ be a Hilbert space of functions on a set $\Omega$
(with the pointwise operations of addition and scalar action).
Its \emph{imbedding constant} $\ccc_{\FFF}$ is defined by
\begin{equation}\label{eq:c}
  \ccc_{\FFF}
  :=
  \sup_{\omega\in\Omega}
  \sup_{f\in\FFF:\left\|f\right\|_{\FFF}\le1}
  f(\omega).
\end{equation}
We will be interested in the case $\ccc_{\FFF}<\infty$
and will refer to $\FFF$ satisfying this condition
as \emph{reproducing kernel Hilbert spaces (RKHS) with finite imbedding constant}.

The Hilbert space $\FFF$ is called a \emph{reproducing kernel Hilbert space} (RKHS)
if all evaluation functionals $f\in\FFF\mapsto f(\omega)$, $\omega\in\Omega$,
are bounded;
the class of RKHS with finite imbedding constant
is a subclass of the class of RKHS.
Let $\FFF$ be an RKHS on $\Omega$.
By the Riesz--Fischer theorem,
for each $\omega\in\Omega$ there exists a function $\kkk_{\omega}\in\FFF$
(the \emph{representer} of $\omega$ in $\FFF$)
such that
\begin{equation}\label{eq:reproducing}
  f(\omega)
  =
  \langle \kkk_{\omega},f\rangle_{\FFF},
  \quad
  \forall f\in\FFF.
\end{equation}
If $\Omega$ is a topological space
and the mapping $\omega\mapsto\kkk_{\omega}$ is continuous,
$\FFF$ is called a \emph{continuous RKHS}.
If $\Omega=\mathbf{X}\times\PPP(\mathbf{Y})\times\mathbf{Y}$
and $\kkk_{\omega}=\kkk_{x,P,y}$ is a continuous function of $(P,y)\in\PPP(\mathbf{Y})\times\mathbf{Y}$
for each $x\in\mathbf{X}$,
we will say that $\FFF$ is \emph{forecast-continuous}.
\begin{theorem}\label{thm:forecasting-bounds}
  Let $\mathbf{Y}$ be a metric compact
  and $\FFF$ be a forecast-continuous RKHS
  on $\mathbf{X}\times\PPP(\mathbf{Y})\times\mathbf{Y}$
  with finite imbedding constant $\ccc_{\FFF}$.
  There is a probability forecasting strategy that guarantees
  \begin{equation*}
    \left|
      \sum_{n=1}^N
      \left(
        f
        \left(
          x_n,P_n,y_n
        \right)
        -
        \int_{\mathbf{Y}}
        f
        \left(
          x_n,P_n,y
        \right)
        P_n(\dd y)
     \right)
    \right|
    \le
    2
    \ccc_{\FFF}
    \left\|
      f
    \right\|_{\FFF}
    \sqrt{N}
  \end{equation*}
  for all $N$ and all $f\in\FFF$.
\end{theorem}

Before proving Theorem \ref{thm:forecasting-bounds}
we will give an example of a convenient RKHS $\FFF$
that can be used in its applications.
Let us consider a finite $\mathbf{Y}$,
represent $\PPP(\mathbf{Y})$ as a simplex in a Euclidean space,
and suppose that $\mathbf{X}$ is a bounded open subset of a Euclidean space.
The interior $\Int\PPP(\mathbf{Y})$ of $\PPP(\mathbf{Y})$
can be regarded as a bounded open subset of a Euclidean space,
and so the product $\mathbf{X}\times\Int\PPP(\mathbf{Y})\times\mathbf{Y}$
can also be regarded as a bounded open set $\Omega$ in a Euclidean space
of dimension $K:=\dim\mathbf{X}+\left|\mathbf{Y}\right|-1$:
namely,
as a disjoint union of $\left|\mathbf{Y}\right|$ copies
of the bounded open set $\mathbf{X}\times\Int\PPP(\mathbf{Y})$.

For a smooth function $u:\Omega\to\bbbr$ and $m\in\{0,1,\ldots\}$ define
\begin{equation}\label{eq:Sobolev}
  \left\|
    u
  \right\|_m
  :=
  \sqrt
  {
    \sum_{0\le\left|\alpha\right|\le m}
    \int_{\Omega}
      \left(
        D^{\alpha}u
      \right)^2
  },
\end{equation}
where $\int_{\Omega}$ stands for the integral with respect to the Lebesgue measure on $\Omega$,
$\alpha$ runs over the multi-indices
$\alpha=(\alpha_1,\ldots,\alpha_K)\in\{0,1,\ldots\}^K$,
and
\begin{equation*}
  \left|
    \alpha
  \right|
  :=
  \alpha_1 + \cdots + \alpha_K,
  \quad
  D^{\alpha}u
  :=
  \frac
  {
     \partial^{\left|\alpha\right|}u
  }
  {
    \partial^{\alpha_1}_{t_1}
    \cdots
    \partial^{\alpha_K}_{t_K}
  }
\end{equation*}
($(t_1,\ldots,t_K)$ is a typical point of the Euclidean space containing $\Omega$).
Let $H^m(\Omega)$ be the completion of the set of smooth function on $\Omega$
with respect to the norm (\ref{eq:Sobolev}).
According to the Sobolev imbedding theorem
(\cite{adams/fournier:2003}, Theorem 4.12),
$H^m(\Omega)$ can be identified with an RKHS of continuous functions
on the closure $\overline{\Omega}$ of $\Omega$
with a finite imbedding constant.
This conclusion depends on the assumption $m>K/2$,
which we will always be making.

It is clear that every continuous function $f$ on $\overline{\Omega}$
can be approximated, arbitrarily closely,
by a function from $H^m(\Omega)$:
even the functions in $C^{\infty}(\bbbr^K)$,
all of which belong to all Sobolev spaces on $\Omega$,
are dense in $C(\overline{\Omega})$
(\cite{adams/fournier:2003}, 2.29).

There is little doubt that Sobolev spaces $H^m(\Omega)$ are continuous
under our assumption $m>K/2$
and for ``nice'' $\Omega$,
although I am not aware of any general results in this direction.

\subsection*{Proof of Theorem \ref{thm:forecasting-bounds}}

If $f:\Omega\to\HHH$ is a function taking values in a topological vector space $\HHH$
and $P$ is a finite measure on its domain $\Omega$,
the integral $\int_{\Omega} f \D P$ will be understood in Pettis's
(\cite{rudin:1991}, Definition 3.26) sense.
Namely,
the integral $\int_{\Omega} f \D P$ is defined to be $h\in\HHH$
such that
\begin{equation}\label{eq:Pettis}
  \Lambda h
  =
  \int_{\Omega} (\Lambda f) \D P
\end{equation}
for all $\Lambda\in\HHH^*$.
The existence and uniqueness of the Pettis integral is assured
if $\Omega$ is a compact topological space
(with $P$ defined on its Borel $\sigma$-algebra),
$\HHH$ is a Banach space, and $f$ is continuous
(\cite{rudin:1991}, Theorems 3.27, 3.20, and 3.3).

\begin{remark}
  {\em Another popular notion of the integral
  for vector-valued functions is Bochner's
  (see, e.g., \cite{yosida:1965}),
  which is more restrictive than Pettis's
  (in particular,
  the Bochner integral always satisfies (\ref{eq:Pettis})).
  Interestingly,
  the Bochner integral $\int_{\Omega} f \D P$
  exists for all measurable functions $f:\Omega\to\HHH$
  (with $\Omega$ a measurable space)
  provided $\HHH$ is a separable Banach space
  and $\int_{\Omega} \left\|f\right\|_{\HHH} \D P<\infty$
  (this follows from Bochner's theorem,
  \cite{yosida:1965}, Theorem 1 in Section V.5,
  and Pettis's measurability theorem,
  \cite{yosida:1965}, the theorem in Section V.4).
  No topological conditions are imposed on $\Omega$ or $f$,
  but there is the requirement of separability
  (which is essential,
  again by Bochner's theorem and Pettis's measurability theorem).
  This requirement, however, may be said to be satisfied automatically
  under the given sufficient conditions for the existence of the Pettis integral:
  since $f(\Omega)$ is a compact metric space,
  it is separable (\cite{engelking:1989}, 4.1.18),
  and we can redefine $\HHH$ as the smallest closed linear subspace
  containing $f(\Omega)$.
  Therefore,
  we can use all properties of the Bochner integral
  under those conditions.}
\end{remark}

We start from a corollary
(a version of Kolmogorov's 1929 result)
of Lemma \ref{lem:takemura}.
\begin{lemma}\label{lem:Hilbert}
  Suppose $\mathbf{Y}$ is a metric compact.
  Let $\Phi_n:\mathbf{X}\times\PPP(\mathbf{Y})\times\mathbf{Y}\to\HHH$,
  $n=1,2,\ldots$,
  be functions taking values in a Hilbert space $\HHH$
  such that, for all $n$ and $x$,
  $\Phi_n(x,P,y)$ is a continuous function of $(P,y)\in\PPP(\mathbf{Y})\times\mathbf{Y}$.
  There is a probability forecasting strategy that guarantees
  \begin{equation}\label{eq:Hilbert}
    \left\|
      \sum_{n=1}^N
      \Psi_n
      \left(
        x_n,P_n,y_n
      \right)
    \right\|_{\HHH}^2
    \le
    \sum_{n=1}^N
    \left\|
      \Psi_n
      \left(
        x_n,P_n,y_n
      \right)
    \right\|_{\HHH}^2
  \end{equation}
  for all $N$,
  where
  \begin{equation*}
    \Psi_n
    \left(
      x,P,y
    \right)
    :=
    \Phi_n
    \left(
      x,P,y
    \right)
    -
    \int_{\mathbf{Y}}
    \Phi_n
    \left(
      x,P,y
    \right)
    P(\dd y).
  \end{equation*}
\end{lemma}
\begin{proof}
  According to Lemma~\ref{lem:takemura},
  it suffices to check that
  \begin{equation}\label{eq:S}
    S_N
    :=
    \left\|
      \sum_{n=1}^N
      \Psi_n
      \left(
        x_n,P_n,y_n
      \right)
    \right\|_{\HHH}^2
    -
    \sum_{n=1}^N
    \left\|
      \Psi_n
      \left(
        x_n,P_n,y_n
      \right)
    \right\|_{\HHH}^2
  \end{equation}
  is the capital process of some strategy for Skeptic
  in the defensive forecasting protocol.
  Since
  \begin{align*}
    S_N - S_{N-1}
    &=
    \left\|
      \sum_{n=1}^{N-1}
      \Psi_n
      \left(
        x_n,P_n,y_n
      \right)
      +
      \Psi_N
      \left(
        x_N,P_N,y_N
      \right)
    \right\|_{\HHH}^2\\
    &\quad{}-
    \left\|
      \sum_{n=1}^{N-1}
      \Psi_n
      \left(
        x_n,P_n,y_n
      \right)
    \right\|_{\HHH}^2
    -
    \left\|
      \Psi_N
      \left(
        x_N,P_N,y_N
      \right)
    \right\|_{\HHH}^2\\
    &=
    \left\langle
      2
      \sum_{n=1}^{N-1}
      \Psi_n
      \left(
        x_n,P_n,y_n
      \right),
      \Psi_N
      \left(
        x_N,P_N,y_N
      \right)
    \right\rangle_{\HHH}\\
    &=
    \left\langle
      A,
      \Psi_N
      \left(
        x_N,P_N,y_N
      \right)
    \right\rangle_{\HHH},
  \end{align*}
  where we have introduced the notation $A$ for the element
  $
    2
    \sum_{n=1}^{N-1}
    \Psi_n
    \left(
      x_n,P_n,y_n
    \right)
  $
  of $\HHH$
  known at the beginning of the $N$th round,
  and, by the definition of the Pettis integral,
  \begin{equation}\label{eq:move}
    \int_{\mathbf{Y}}
    \left\langle
      A,
      \Psi_N
      \left(
        x_N,P_N,y
      \right)
    \right\rangle_{\HHH}
    P_N(\dd y)
    =
    \left\langle
      A,
      \int_{\mathbf{Y}}
      \Psi_N
      \left(
        x_N,P_N,y
      \right)
      P_N(\dd y)
    \right\rangle_{\HHH}
    =
    0,
  \end{equation}
  the difference $S_N-S_{N-1}$ coincides with Skeptic's gain
  in the $N$th round of the testing protocol
  when he makes the valid move
  $
    f_N(y)
    :=
    \left\langle
      A,
      \Psi_N
      \left(
        x_N,P_N,y
      \right)
    \right\rangle_{\HHH}
  $.
  It remains to check that
  $
    F_N(y,P)
    :=
    \left\langle
      A,
      \Psi_N
      \left(
        x_N,P,y
      \right)
    \right\rangle_{\HHH}
  $
  will be a valid move in the defensive forecasting protocol,
  i.e., that the function $F_N$ is lower semicontinuous;
  we will see that it is in fact continuous.
  By Lemma \ref{lem:continuity1} below,
  the function $\int_{\mathbf{Y}}\Phi_N(x,P,y)P(\dd y)$ is continuous in $P$;
  therefore,
  the function $\Psi_N$ is continuous in $(P,y)$.
  This implies that
  $
    \left\langle
      A,
      \Psi_N
      \left(
        x_N,P,y
      \right)
    \right\rangle_{\HHH}
  $
  is a continuous function of $(P,y)$.
  \qedtext
\end{proof}
The proof of Lemma \ref{lem:Hilbert} used the following lemma.
\begin{lemma}\label{lem:continuity1}
  Suppose $\mathbf{Y}$ is a metric compact
  and $\Phi:\PPP(\mathbf{Y})\times\mathbf{Y}\to\HHH$
  is a continuous mapping into a Hilbert space $\HHH$.
  The mapping
  $P\in\PPP(\mathbf{Y})\mapsto\int_{\mathbf{Y}}\Phi(P,y)P(\dd y)$
  is also continuous.
\end{lemma}
\begin{proof}
  Let $P_n\to P$ as $n\to\infty$;
  our goal is to prove that
  $\int_{\mathbf{Y}}\Phi(P_n,y)P_n(\dd y)\to\int_{\mathbf{Y}}\Phi(P,y)P(\dd y)$.
  We have:
  \begin{multline}\label{eq:decomposition}
    \left\|
      \int_{\mathbf{Y}}\Phi(P_n,y)P_n(\dd y)
      -
      \int_{\mathbf{Y}}\Phi(P,y)P(\dd y)
    \right\|_{\HHH}\\
    \le
    \left\|
      \int_{\mathbf{Y}}\Phi(P_n,y)P_n(\dd y)
      -
      \int_{\mathbf{Y}}\Phi(P,y)P_n(\dd y)
    \right\|_{\HHH}\\
    +
    \left\|
      \int_{\mathbf{Y}}\Phi(P,y)P_n(\dd y)
      -
      \int_{\mathbf{Y}}\Phi(P,y)P(\dd y)
    \right\|_{\HHH}.
  \end{multline}
  The first addend on the right-hand side can be bounded above by
  \begin{equation*}
    \int_{\mathbf{Y}}
      \left\|
        \Phi(P_n,y)
        -
        \Phi(P,y)
      \right\|_{\HHH}
    P_n(\dd y)
  \end{equation*}
  (\cite{rudin:1991}, 3.29),
  and the last expression tends to zero
  since $\Phi$ is uniformly continuous
  (\cite{engelking:1989}, 4.3.32).
  The second addend on the right-hand side of (\ref{eq:decomposition})
  tends to zero
  by the continuity of the mapping $Q\in\PPP(\mathbf{Y})\mapsto\int_{\mathbf{Y}}f(y)Q(\dd y)$
  for a continuous $f$
  (\cite{bourbaki:integration}, III.4.2, Proposition 6).
  \qedtext
\end{proof}
The following variation on Lemma \ref{lem:continuity1}
will be needed later.
\begin{lemma}\label{lem:continuity2}
  Suppose $\mathbf{X}$ and $\mathbf{Y}$ are metric compacts
  and $\Phi:\mathbf{X}\times\PPP(\mathbf{Y})\times\mathbf{Y}\to\HHH$
  is a continuous mapping into a Hilbert space $\HHH$.
  The mapping
  $(x,P)\in\mathbf{X}\times\PPP(\mathbf{Y})\mapsto\int_{\mathbf{Y}}\Phi(x,P,y)P(\dd y)$
  is also continuous.
\end{lemma}
\begin{proof}
  Let $x_n\to x$ and $P_n\to P$ as $n\to\infty$.
  To prove
  $\int_{\mathbf{Y}}\Phi(x_n,P_n,y)P_n(\dd y)\to\int_{\mathbf{Y}}\Phi(x,P,y)P(\dd y)$
  we can use a similar argument to that in the previous lemma
  applied to
  \begin{multline*}
    \left\|
      \int_{\mathbf{Y}}\Phi(x_n,P_n,y)P_n(\dd y)
      -
      \int_{\mathbf{Y}}\Phi(x,P,y)P(\dd y)
    \right\|_{\HHH}\\
    \le
    \left\|
      \int_{\mathbf{Y}}\Phi(x_n,P_n,y)P_n(\dd y)
      -
      \int_{\mathbf{Y}}\Phi(x,P,y)P_n(\dd y)
    \right\|_{\HHH}\\
    +
    \left\|
      \int_{\mathbf{Y}}\Phi(x,P,y)P_n(\dd y)
      -
      \int_{\mathbf{Y}}\Phi(x,P,y)P(\dd y)
    \right\|_{\HHH}.
    \qedmath 
  \end{multline*}
\end{proof}

Now we can begin the actual proof of Theorem \ref{thm:forecasting-bounds}.
Take as $\Phi(x,P,y)$ the representer $\kkk_{x,P,y}$
of the evaluation functional $f\in\FFF\mapsto f(x,P,y)$:
\begin{equation*}
  \left\langle
    f,
    \kkk_{x,P,y}
  \right\rangle_{\FFF}
  =
  f(x,P,y),
  \quad
  \forall (x,P,y) \in \mathbf{X}\times\PPP(\mathbf{Y})\times\mathbf{Y},
  f\in\FFF.
\end{equation*}
Set
\begin{equation*}
  \kkk_{x,P}
  :=
  \int_{\mathbf{Y}}
  \kkk_{x,P,y}
  P(\dd y);
\end{equation*}
the function $\kkk_{x,P}$ is continuous in $P$ by Lemma \ref{lem:continuity1}.

Theorem \ref{thm:forecasting-bounds} will easily follow from the following lemma,
which itself is an easy implication of Lemma \ref{lem:Hilbert}.
\begin{lemma}\label{lem:RKHS}
  Let $\mathbf{Y}$ be a metric compact
  and $\FFF$ be a forecast-continuous RKHS
  on $\mathbf{X}\times\PPP(\mathbf{Y})\times\mathbf{Y}$.
  There is a probability forecasting strategy that guarantees
  \begin{multline*}
    \left|
      \sum_{n=1}^N
      \left(
        f
        \left(
          x_n,P_n,y_n
        \right)
        -
        \int_{\mathbf{Y}}
        f
        \left(
          x_n,P_n,y
        \right)
        P_n(\dd y)
     \right)
    \right|\\
    \le
    \left\|
      f
    \right\|_{\FFF}
    \sqrt
    {
      \sum_{n=1}^N
      \left\|
        \kkk_{x_n,P_n,y_n}
        -
        \kkk_{x_n,P_n}
      \right\|_{\FFF}^2
    }
  \end{multline*}
  for all $N$ and all $f\in\FFF$.
\end{lemma}
\begin{proof}
  Using Lemma \ref{lem:Hilbert}
  (with all $\Psi_n$ equal,
  $\Psi_n(x,P,y):=\kkk_{x,P,y}-\kkk_{x,P}$),
  we obtain:
  \begin{multline*}
    \left|
      \sum_{n=1}^N
      \left(
        f
        \left(
          x_n,P_n,y_n
        \right)
        -
        \int_{\mathbf{Y}}
        f
        \left(
          x_n,P_n,y
        \right)
        P_n(\dd y)
     \right)
    \right|\\
    =
    \left|
      \sum_{n=1}^N
      \left(
        \left\langle
          f,
          \kkk_{x_n,P_n,y_n}
        \right\rangle_{\FFF}
        -
        \int_{\mathbf{Y}}
        \left\langle
          f,
          \kkk_{x_n,P_n,y}
        \right\rangle_{\FFF}
        P_n(\dd y)
     \right)
    \right|\\
    =
    \left|
      \left\langle
        f,
        \sum_{n=1}^N
        \left(
          \kkk_{x_n,P_n,y_n}
          -
          \kkk_{x_n,P_n}
        \right)
      \right\rangle_{\FFF}
    \right|
    \le
    \left\|
      f
    \right\|_{\FFF}
    \left\|
      \sum_{n=1}^N
      \left(
        \kkk_{x_n,P_n,y_n}
        -
        \kkk_{x_n,P_n}
      \right)
    \right\|_{\FFF}\\
    \le
    \left\|
      f
    \right\|_{\FFF}
    \sqrt
    {
      \sum_{n=1}^N
      \left\|
        \kkk_{x_n,P_n,y_n}
        -
        \kkk_{x_n,P_n}
      \right\|_{\FFF}^2
    }.
    \qedmath
  \end{multline*}
\end{proof}

\begin{remark}
  {\em The algorithm of Lemma \ref{lem:RKHS} is a generalization of the K29 algorithm
  of \cite{\GTPVIII}.
  It would be interesting also to analyze the K29${}^*$ algorithm
  (called the algorithm of large numbers in \cite{\GTPXIII} and \cite{\GTPXIV}).}
\end{remark}

To deduce Theorem \ref{thm:forecasting-bounds}
from Lemma \ref{lem:RKHS},
notice that
$\left\|\kkk_{x,P,y}\right\|_{\FFF}\le\ccc_{\FFF}$
(by Lemma \ref{lem:optimization} below),
$\left\|\kkk_{x,P}\right\|_{\FFF}
\le\int_{\mathbf{Y}}\left\|\kkk_{x,P,y}\right\|_{\FFF}P(\dd y)
\le\ccc_{\FFF}$,
and, therefore,
\begin{equation*}
  \sum_{n=1}^N
  \left\|
    \kkk_{x_n,P_n,y_n}
    -
    \kkk_{x_n,P_n}
  \right\|_{\FFF}^2
  \le
  4\ccc_{\FFF}^2N.
\end{equation*}
This completes the proof apart from Lemma \ref{lem:optimization}.

Let $\FFF$ be an RKHS on $\Omega$.
The norm of the evaluation functional $f\in\FFF\mapsto f(\omega)$
will be denoted by $\ccc_{\FFF}(\omega)$.
It is clear that $\FFF$ is an RKHS with finite imbedding constant
if and only if
\begin{equation}\label{eq:c-mod}
  \ccc_{\FFF}
  :=
  \sup_{\omega\in\Omega}
  \ccc_{\FFF}(\omega)
\end{equation}
is finite;
the constants in (\ref{eq:c-mod}) and (\ref{eq:c}) coincide.
The next lemma,
concluding the proof of Theorem \ref{thm:forecasting-bounds},
asserts that the norm $\left\|\kkk_{\omega}\right\|_{\FFF}$
of the representer of $\omega$ in $\FFF$
coincides with the norm $\ccc_{\FFF}(\omega)$ of the evaluation functional $f\mapsto f(\omega)$.
\begin{lemma}\label{lem:optimization}
  Let $\FFF$ be an RKHS on $\Omega$.
  For each $\omega\in\Omega$,
  \begin{equation}\label{eq:equal}
    \left\|\kkk_{\omega}\right\|_{\FFF}
    =
    \ccc_{\FFF}(\omega).
  \end{equation}
\end{lemma}
\begin{proof}
  Fix $\omega\in\Omega$.
  We are required to prove
  \begin{equation*}
    \sup_{f:\left\|f\right\|_{\FFF}\le1}
    \left|
      f(\omega)
    \right|
    =
    \left\|\kkk_{\omega}\right\|_{\FFF}.
  \end{equation*}
  The inequality $\le$ follows from
  \begin{equation*}
    \left|
      f(\omega)
    \right|
    =
    \left|
      \left\langle
        f,\kkk_{\omega}
      \right\rangle_{\FFF}
    \right|
    \le
    \left\|
      f
    \right\|_{\FFF}
    \left\|
      \kkk_{\omega}
    \right\|_{\FFF}
    \le
    \left\|
      \kkk_{\omega}
    \right\|_{\FFF},
  \end{equation*}
  where $\left\|f\right\|_{\FFF}\le1$.
  The inequality $\ge$ follows from
  \begin{equation*}
    \left|
      f(\omega)
    \right|
    =
    \frac
    {\kkk_{\omega}(\omega)}
    {
      \left\|
        \kkk_{\omega}
      \right\|_{\FFF}
    }
    =
    \frac
    {
      \left\langle
        \kkk_{\omega},\kkk_{\omega}
      \right\rangle_{\FFF}
    }
    {
      \left\|
        \kkk_{\omega}
      \right\|_{\FFF}
    }
    =
    \left\|
      \kkk_{\omega}
    \right\|_{\FFF},
  \end{equation*}
  where $f:=\kkk_{\omega}/\left\|\kkk_{\omega}\right\|_{\FFF}$
  and $\left\|\kkk_{\omega}\right\|_{\FFF}$ is assumed to be non-zero
  (if it is zero, $\kkk_{\omega}=0$, which implies $\ccc_{\FFF}(\omega)=0$,
  and (\ref{eq:equal}) still holds).
  \qedtext
\end{proof}

\subsection*{Reproducing kernels}

In this subsection we start preparations for proving Theorem \ref{thm:forecasting-asymptotic}.
But first we need to delve slightly deeper into the theory of RKHS.
An equivalent language for talking about RKHS
is provided by the notion of a reproducing kernel,
and this subsection defines reproducing kernels
and summarizes some of their properties.
For a detailed discussion,
see, e.g., \cite{aronszajn:1944,aronszajn:1950} or \cite{meschkowski:1962}.

The \emph{reproducing kernel} of an RKHS $\FFF$ on $\Omega$
is the function $\kkk:\Omega^2\to\bbbr$ defined by
\begin{equation*}
  \kkk(\omega,\omega')
  :=
  \left\langle
    \kkk_{\omega},\kkk_{\omega'}
  \right\rangle_{\FFF}
\end{equation*}
(equivalently, we could define $\kkk(\omega,\omega')$ as $\kkk_{\omega}(\omega')$
or as $\kkk_{\omega'}(\omega)$).
The origin of this name is the ``reproducing property'' (\ref{eq:reproducing}).

There is a simple internal characterization of reproducing kernels of RKHS.
First,
it is easy to check that the function $\kkk(\omega,\omega')$,
as we defined it,
is symmetric,
\begin{equation*}
  \kkk(\omega,\omega')=\kkk(\omega',\omega),
  \quad
  \forall (\omega,\omega')\in \Omega^2,
\end{equation*}
and positive definite,
\begin{multline*}
  \sum_{i=1}^m\sum_{j=1}^m t_i t_j \kkk(\omega_i,\omega_j)\ge0,\\
  \forall m=1,2,\ldots,
  (t_1,\ldots,t_m)\in\bbbr^m,
  (\omega_1,\dots,\omega_m)\in \Omega^m.
\end{multline*}
On the other hand,
for every symmetric and positive definite $\kkk:\Omega^2\to\bbbr$
there exists a unique RKHS $\FFF$ on $\Omega$
such that $\kkk$ is the reproducing kernel of $\FFF$
(\cite{aronszajn:1944}, Theorem 2 on p.~143).

We can see that the notions of a reproducing kernel of RKHS
and of a symmetric positive definite function on $\Omega^2$
have the same content,
and we will sometimes say ``kernel on $\Omega$''
to mean a symmetric positive definite function
on $\Omega^2$.
Kernels in this sense are the main source of RKHS in learning theory:
cf.\ \cite{vapnik:1998,scholkopf/smola:2002,shawe-taylor/cristianini:2004}.
Every kernel on $\mathbf{X}$ is a valid parameter
for our prediction algorithms.
In general,
it is convenient to use RKHS in stating mathematical properties
of prediction algorithms,
but the algorithms themselves typically use the more constructive representation of RKHS
via their reproducing kernels.

It is easy to see
that $\FFF$ is a continuous RKHS if and only if its reproducing kernel is continuous
(see \cite{steinwart:2001} or \cite{\GTPXIII}, Appendix B of the arXiv technical report).
A convenient equivalent definition of $\ccc_{\FFF}$ is
\begin{equation}\label{eq:equiv}
  \ccc_{\FFF}
  =
  \ccc_{\kkk}
  :=
  \sup_{\omega\in \Omega}
  \sqrt{\kkk(\omega,\omega)}
  =
  \sup_{\omega,\omega'\in \Omega}
  \sqrt{\left|\kkk(\omega,\omega')\right|},
\end{equation}
$\kkk$ being the reproducing kernel of an RKHS $\FFF$ on $\Omega$.

Let us say that a family $\FFF$ of functions $f:\Omega\to\bbbr$ is \emph{universal}
if $\Omega$ is a topological space
and for every compact subset $A$ of $\Omega$
every continuous function on $A$
can be arbitrarily well approximated in the metric $C(A)$
by functions in $\FFF$
(in the case of compact $\Omega$ this coincides
with the definition given in \cite{steinwart:2001} as Definition 4).

We have already noticed the obvious fact
that the Sobolev spaces $H^m(\Omega)$ on bounded open $\Omega\subseteq\bbbr^K$,
$K<2m$, are universal.
There is a price to pay for the obviousness of this fact:
the reproducing kernels of the Sobolev spaces
are known only in some special cases
(see, e.g., \cite{berlinet/thomas-agnan:2004}, Section 7.4).
This complicates checking their continuity.

On the other hand,
some very simple continuous reproducing kernels,
such as the Gaussian kernel
\begin{equation*}
  \kkk(\omega,\omega')
  :=
  \exp
  \left(
    -\frac{\left\|\omega-\omega'\right\|^2}{\sigma^2}
  \right)
\end{equation*}
($\left\|\cdot\right\|$ being the Euclidean norm and $\sigma$ being an arbitrary positive constant)
on the Euclidean space $\bbbr^K$
and the infinite polynomial kernel
\begin{equation*}
  \kkk(\omega,\omega')
  :=
  \frac{1}{1-\langle \omega,\omega'\rangle}
\end{equation*}
($\langle\cdot,\cdot\rangle$ being the Euclidean inner product)
on the Euclidean ball $\{\omega\in\bbbr^K\st\|\omega\|<1\}$,
are universal
(\cite{steinwart:2001}, Examples 1 and 2).
Their universality is not difficult to prove
but not obvious
(and even somewhat counterintuitive in the case of the Gaussian kernel:
\emph{a priori} one might expect
that only smooth functions
that are almost linear at scales smaller than $\sigma$
can belong to the corresponding RKHS).
On the other hand, their continuity is obvious.

\subsection*{Universal function space on the Hilbert cube}

Remember that the \emph{Hilbert cube} is the topological space $[0,1]^{\infty}$
(\cite{engelking:1989}, 2.3.22),
i.e.,
the topological product of a countable number of closed intervals $[0,1]$.
As the next step in the proof of Theorem \ref{thm:forecasting-asymptotic},
in this subsection
we construct a universal RKHS on the Hilbert cube with finite imbedding constant;
the idea of the construction is to ``mix'' Sobolev spaces on $[0,1]^K$
for $K=1,2,\ldots$
(or the spaces mentioned at the end of the previous subsection,
for which both continuity and universality are proven).

Let $\FFF_K$, $K=1,2,\ldots$,
be the set of all functions $f$ on the Hilbert cube
such that $f(t_1,t_2,\ldots)$ depends only on $t_1,\ldots,t_K$
and whose norm (\ref{eq:Sobolev}) (with $\Omega:=[0,1]^K$)
is finite for $m:=K$.
Equipping $\FFF_K$ with this norm
we obtain an RKHS with finite imbedding constant.
Let $c_K$ be the imbedding constant of $\FFF_K$.
It will be convenient to modify each $\FFF_K$ by scaling the inner product:
\begin{equation*}
  \left\langle
    \cdot,\cdot
  \right\rangle_{\FFF'_K}
  :=
  c_K^2 2^{K}
  \left\langle
    \cdot,\cdot
  \right\rangle_{\FFF_K};
\end{equation*}
the scaled $\FFF_K$ will be denoted $\FFF'_K$.
By (\ref{eq:reproducing}),
the representer $\kkk'_{\omega}$ of $\omega$ in $\FFF'_K$ can be expressed as
$\kkk'_{\omega} = c_K^{-2} 2^{-K} \kkk_{\omega}$
via the representer $\kkk_{\omega}$ of $\omega$ in $\FFF_K$.
Therefore, the imbedding constant of $\FFF'_K$ is $2^{-K/2}$,
and it is obvious that $\FFF'_K$ inherits from $\FFF_K$ the property of being
a universal RKHS for functions that only depend on $t_1,\ldots,t_K$.

For the reproducing kernel $\kkk'_K(\omega,\omega')$ of $\FFF'_K$
we have
\begin{equation*}
  \left|
    \kkk'_K(\omega,\omega')
  \right|
  =
  \left|
    \left\langle
      \kkk'_{\omega},\kkk'_{\omega'}
    \right\rangle_{\FFF'_K}
  \right|
  \le
  \left\|
    \kkk'_{\omega}
  \right\|_{\FFF'_K}
  \left\|
    \kkk'_{\omega'}
  \right\|_{\FFF'_K}
  \le
  2^{-K/2}
  2^{-K/2}
  =
  2^{-K},
\end{equation*}
where $\kkk'_{\omega}$ and $\kkk'_{\omega'}$ stand for the representers in $\FFF'_K$.
Define an RKHS $\GGG_K$ as the set of all functions
$f:[0,1]^{\infty}\to\bbbr$
that can be decomposed into a sum $f=f_1+\cdots+f_K$,
where $f_k\in\FFF'_k$, $k=1,\ldots,K$.
The norm of $f$ is defined as the infimum
\begin{equation*}
  \left\|
    f
  \right\|_{\GGG_K}
  :=
  \inf
  \sqrt
  {
    \sum_{k=1}^K
    \left\|
      f_k
    \right\|^2_{\FFF'_k}
  }
\end{equation*}
over all such decompositions.
According to the theorem on p.~353 of \cite{aronszajn:1950},
$\GGG_K$ is an RKHS whose reproducing kernel $\kkk_K$ satisfies
\begin{equation*}
  \kkk_K(\omega,\omega')
  =
  \sum_{k=1}^K
  \kkk'_k(\omega,\omega')
  \in
  \left[
    -1+2^{-K},
    1-2^{-K}
  \right].
\end{equation*}

The limiting RKHS of $\GGG_K$, $K\to\infty$,
is defined in \cite{aronszajn:1950}, Section I.9 (Case B),
in two steps.
Let $\FFF_0$ consist of the functions in $\GGG_K$, $K=1,2,\ldots$;
the $\FFF_0$-norm of a function $g\in\GGG_K$ is defined as
\begin{equation*}
  \left\|
    g
  \right\|_{\FFF_0}
  :=
  \inf_{k\ge K}
  \left\|
    g
  \right\|_{\GGG_k}.
\end{equation*}
In general, the space $\FFF_0$ is not complete.
Therefore,
a larger space $\FFF_0^*$ is defined:
$f\in\FFF_0^*$ if there is a Cauchy sequence $f_n$ in $\FFF_0$ such that
\begin{equation}\label{eq:aronszajn15}
  \forall\omega\in[0,1]^{\infty}:
  f(\omega)
  =
  \lim_{n\to\infty}
  f_n(\omega);
\end{equation}
the norm of such an $f$ is defined as
\begin{equation*}
  \left\|
    f
  \right\|_{\FFF_0^*}
  :=
  \inf
  \lim_{n\to\infty}
  \left\|
    f_n
  \right\|_{\FFF_0},
\end{equation*}
where the infimum is taken over all Cauchy sequences
satisfying (\ref{eq:aronszajn15}).
By Theorem II on p.~367 of \cite{aronszajn:1950},
$\FFF_0^*$ is an RKHS with reproducing kernel
\begin{equation}\label{eq:universal-kernel}
  \kkk^*(\omega,\omega')
  =
  \sum_{k=1}^{\infty}
  \kkk'_k(\omega,\omega')
  \in
  [-1,1];
\end{equation}
therefore, its imbedding constant is finite
(at most 1: see (\ref{eq:equiv})).

\begin{lemma}
  The RKHS $\FFF_0^*$ on the Hilbert cube is universal and continuous.
\end{lemma}
\begin{proof}
  The Hilbert cube is a topological space
  that is both compact
  (by Tikhonov's theorem,
  \cite{engelking:1989}, 3.2.4)
  and metrizable;
  for concreteness,
  let us fix the metric
  \begin{equation*}
    \rho
    \left(
      \left(
        t_1,t_2,\ldots
      \right),
      \left(
        t'_1,t'_2,\ldots
      \right)
    \right)
    :=
    \sum_{k=1}^{\infty}
    2^{-k}
    \left|
      t_k-t'_k
    \right|.
  \end{equation*}
  Let $f$ be a continuous function on the Hilbert cube.
  Since every continuous function on a compact metric space is uniformly continuous
  (\cite{engelking:1989}, 4.3.32),
  the function
  \begin{equation*}
    g
    \left(
      t_1,t_2,\ldots
    \right)
    :=
    f
    \left(
      t_1,\ldots,t_K,0,0,\ldots
    \right)
  \end{equation*}
  can be made arbitrarily close to $f$, in metric $C([0,1]^{\infty})$,
  by making $K$ sufficiently large.
  It remains to notice that $g$ can be arbitrarily closely approximated
  by a function in $\FFF_K$
  and that every function in $\FFF_K$ belongs to $\FFF_0^*$.

  The continuity of $\FFF_0^*$ follows
  from the Weierstrass $M$-test
  and the expression (\ref{eq:universal-kernel})
  of its reproducing kernel
  via the reproducing kernels of the spaces $\FFF'_K$, $K=1,2,\ldots$,
  with imbedding constant $2^{-K}$.
  \qedtext
\end{proof}
\begin{corollary}\label{cor:universal-compact}
  For any compact metric space $\Omega$
  there is a continuous universal RKHS $\FFF$ on $\Omega$
  with finite imbedding constant.
\end{corollary}
\begin{proof}
  It is known
  (\cite{engelking:1989}, 4.2.10)
  that every compact metric space can be homeo\-morphically imbedded into the Hilbert cube;
  let $F:\Omega\to[0,1]^{\infty}$ be such an imbedding.
  The image $F(\Omega)$ is a compact subset of the Hilbert cube
  (\cite{engelking:1989}, 3.1.10).
  Let $\FFF$ be the class of all functions $f:\Omega\to\bbbr$
  such that $f(F^{-1}):F(\Omega)\to\bbbr$
  is the restriction of a function in $\FFF_0^*$ to $F(\Omega)$;
  the norm of $f$ is defined as the infimum of the norms of the extensions of $f(F^{-1})$
  to the whole of the Hilbert cube.
  According to the theorem on p.~351 of \cite{aronszajn:1950},
  this function space is an RKHS
  whose reproducing kernel is
  $\kkk(\omega,\omega'):=\kkk^*(F(\omega),F(\omega'))$,
  where $\kkk^*$ is the reproducing kernel of $\FFF_0^*$;
  we can see that $\FFF$ is a continuous RKHS with finite imbedding constant.

  Let us see that the RKHS $\FFF$ is universal.
  Take any continuous function $g:\Omega\to\bbbr$.
  By the Tietze--Uryson theorem (\cite{engelking:1989}, 2.1.8),
  $g(F^{-1}):F(\Omega)\to\bbbr$
  can be extended to a continuous function $g_1$ on $[0,1]^{\infty}$.
  Let $g_2\in\FFF_0^*$ be a function that is close to $g_1$
  in the $C([0,1]^{\infty})$ norm.
  Then $g_2(F):\Omega\to\bbbr$ will belong to $\FFF$
  and will be close to $g$ in the $C(\Omega)$ norm.
  \qedtext
\end{proof}

\subsection*{Proof of Theorem \ref{thm:forecasting-asymptotic}}

We start by proving the theorem
under the assumption that $\mathbf{X}$ and $\mathbf{Y}$
are compact metric spaces.
As explained above,
in this case $\PPP(\mathbf{Y})$ is also compact and metrizable;
therefore,
$\Omega:=\mathbf{X}\times\PPP(\mathbf{Y})\times\mathbf{Y}$
is also compact
and metrizable.
Let $f$ be a continuous real-valued function on $\Omega$;
our goal is to establish the consequent of (\ref{eq:calibration-cum-resolution}).

Let $\FFF$ be a universal and continuous RKHS on $\Omega$
with finite imbedding constant
(cf.\ Corollary \ref{cor:universal-compact}).
If $g\in\FFF$ is at a distance at most $\epsilon$ from $f$
in the $C(\Omega)$ metric,
we obtain from Theorem \ref{thm:forecasting-bounds}:
\begin{multline}\label{eq:approximation}
  \limsup_{N\to\infty}
  \left|
    \frac1N
    \sum_{n=1}^N
    \left(
      f
      \left(
        x_n,P_n,y_n
      \right)
      -
      \int_{\mathbf{Y}}
      f
      \left(
        x_n,P_n,y
      \right)
      P_n(\dd y)
   \right)
  \right|\\
  \le
  \limsup_{N\to\infty}
  \left|
    \frac1N
    \sum_{n=1}^N
    \left(
      g
      \left(
        x_n,P_n,y_n
      \right)
      -
      \int_{\mathbf{Y}}
      g
      \left(
        x_n,P_n,y
      \right)
      P_n(\dd y)
   \right)
  \right|
  +
  2\epsilon
  =
  2\epsilon.
\end{multline}
Since this can be done for any $\epsilon>0$,
the proof for the case of compact $\mathbf{X}$ and $\mathbf{Y}$ is complete.

The rest of the proof is based on the following game
(an abstract version of the ``doubling trick'',
\cite{cesabianchi/lugosi:2006})
played in a topological space $X$:

\bigskip

\noindent
\textsc{Game of removal $G(X)$}\nopagebreak
\begin{tabbing}
  \qquad\=\qquad\=\qquad\kill
  FOR $n=1,2,\dots$:\\
  \> Remover announces compact $K_n\subseteq X$.\\
  \> Evader announces $p_n\notin K_n$.\\
  END FOR.
\end{tabbing}
\textbf{Winner:}
Evader if the set $\left\{p_1,p_2,\ldots\right\}$ is precompact;
Remover otherwise.

\bigskip

\noindent
Intuitively,
the goal of Evader is to avoid being removed to the infinity.
Without loss of generality
we will assume that Remover always announces a non-decreasing sequence of compact sets:
$K_1\subseteq K_2\subseteq\cdots$.
\begin{lemma}[Gruenhage]\label{lem:Gruenhage}
  Remover has a winning strategy in $G(X)$
  if $X$ is a locally compact and paracompact space.
\end{lemma}
\begin{proof}
  We will follow the proof of Theorem 4.1 in \cite{gruenhage:2006}
  (the easy direction).
  If $X$ is locally compact and $\sigma$-compact,
  there exists a non-decreasing sequence $K_1\subseteq K_2\subseteq\cdots$
  of compact sets covering $X$,
  and each $K_n$ can be extended to compact $K^*_n$
  so that $\Int K^*_n\supseteq K_n$
  (\cite{engelking:1989}, 3.3.2).
  Remover will obviously win $G(X)$ choosing $K^*_1,K^*_2,\ldots$ as his moves.

  If $X$ is the sum of locally compact $\sigma$-compact spaces $X_s$, $s\in S$,
  Remover plays, for each $s\in S$, the strategy described in the previous paragraph
  on the subsequence of Evader's moves belonging to $X_s$.
  If Evader chooses $p_n\in X_s$ for infinitely many $X_s$,
  those $X_s$ will form an open cover of the closure of $\{p_1,p_2,\ldots\}$
  without a finite subcover.
  If $x_n$ are chosen from only finitely many $X_s$,
  there will be infinitely many $x_n$ chosen from some $X_s$,
  and the result of the previous paragraph can be applied.
  It remains to remember that each locally compact paracompact
  can be represented as the sum of locally compact $\sigma$-compact subsets
  (\cite{engelking:1989}, 5.1.27).
  \qedtext
\end{proof}

Now it is easy to prove the general theorem\label{p:general}.
Forecaster's strategy ensuring (\ref{eq:calibration-cum-resolution})
will be constructed from his strategies $\SSS(A,B)$
ensuring the consequent of (\ref{eq:calibration-cum-resolution})
under the condition $\forall n:(x_n,y_n)\in A\times B$
for given compact sets $A\subseteq\mathbf{X}$ and $B\subseteq\mathbf{Y}$
and from Remover's winning strategy in $G(\mathbf{X}\times\mathbf{Y})$
(remember that, by Stone's theorem, \cite{engelking:1989}, 5.1.3,
all metric space are paracompact
and that the product of two locally compact spaces is locally compact,
\cite{engelking:1989}, 3.3.13;
therefore, Lemma \ref{lem:Gruenhage} is applicable to $G(\mathbf{X}\times\mathbf{Y})$).
Without loss of generality we assume that Remover's moves
are always of the form $A\times B$
for $A\subseteq\mathbf{X}$ and $B\subseteq\mathbf{Y}$.
Forecaster will be playing two games in parallel:
the probability forecasting game
and the auxiliary game of removal $G(\mathbf{X}\times\mathbf{Y})$
(in the role of Evader).

Forecaster asks Remover to make his first move $A_1\times B_1$ in the game of removal.
He then plays the probability forecasting game
using the strategy $\SSS(A_1,B_1)$
until Reality chooses $(x_n,y_n)\notin A_1\times B_1$
(forever if Reality never chooses such $(x_n,y_n)$).
As soon as such $(x_n,y_n)$ is chosen,
Forecaster, in his Evader hat,
announces $(x_n,y_n)$ and notes Remover's move $(A_2,B_2)$.
He then plays the probability forecasting game
using the strategy $\SSS(A_2,B_2)$
until Reality chooses $(x_n,y_n)\notin A_2\times B_2$,
etc.

Let us check that this strategy for Forecaster
will always ensure (\ref{eq:calibration-cum-resolution}).
If Reality chooses $(x_n,y_n)$ outside Forecaster's current $A_k\times B_k$
finitely often,
the consequent of (\ref{eq:calibration-cum-resolution}) will be satisfied.
If Reality chooses $(x_n,y_n)$ outside Forecaster's current $A_k\times B_k$
infinitely often,
the set $\{(x_n,y_n)\st n=1,2,\ldots\}$ will not be precompact,
and so the antecedent of (\ref{eq:calibration-cum-resolution}) will be violated.

\section{Implications for probability theory}

This section is an aside;
its results are not used in the rest of the paper.

As we discussed at the end of Section \ref{sec:S},
the procedure of defensive forecasting can be applied to virtually any law of probability
(stated game-theoretically)
to obtain a probability forecasting strategy
whose forecasts are guaranteed to satisfy this law.
Unfortunately, the standard laws of probability theory
are often not strong enough to produce interesting probability forecasting strategies
(\cite{\GTPVIII}, Section 4.1).
In particular, for the purpose of this paper it would be easiest
to apply the procedure of defensive forecasting
to a law of probability asserting that (\ref{eq:calibration-cum-resolution})
holds for all continuous functions $f$ simultaneously
with probability one.
I am not aware of such results,
but in the derivation of Theorem \ref{thm:true-asymptotic}
we essentially proved one.
In this section this result will be stated formally
(as Theorem \ref{thm:true-asymptotic}).

In general,
it can be hoped that probability theory and competitive on-line prediction
have a potential to enrich each other;
not only laws of probability can be translated into probability forecasting strategies
via defensive forecasting,
but also the needs of competitive on-line prediction
can help identify and fill gaps in the existing probability theory.

\subsection*{Game-theoretic result}

Let us say that Skeptic \emph{can force} some property $E$ of the players' moves
$x_n,P_n,y_n$, $n=1,2,\ldots$,
in the testing protocol if he has a strategy guaranteeing that
(1) his capital $\K_n$ is always non-negative, and
(2) either $E$ is satisfied or $\lim_{n\to\infty}\K_n=\infty$.
The properties that can be forced by Skeptic
are the game-theoretic analogue of the properties that hold with probability one
in measure-theoretic probability theory
(\cite{shafer/vovk:2001}, Section 8.1).

The following is a corollary from the proof
(rather than the statement,
which is why we also call it a theorem)
of Theorem \ref{thm:forecasting-asymptotic}.
Its interpretation is that the true probabilities have good calibration-cum-resolution.
\begin{theorem}\label{thm:true-asymptotic}
  Suppose $\mathbf{X}$ and $\mathbf{Y}$ are locally compact metric spaces.
  Skeptic can force
  \begin{multline}\label{eq:true-calibration-cum-resolution}
    \left(
      \left\{
        x_1,x_2,\ldots
      \right\}
      \text{ and }
      \left\{
        y_1,y_2,\ldots
      \right\}
      \text{ are precompact}
    \right)
    \Longrightarrow{}\\
    \left(
      \forall f:
      \lim_{N\to\infty}
      \frac1N
      \sum_{n=1}^N
      \left(
        f
        \left(
          x_n,P_n,y_n
        \right)
        -
        \int_{\mathbf{Y}}
          f
        \left(
          x_n,P_n,y
        \right)
        P_n(\dd y)
      \right)
      =
      0
    \right)
  \end{multline}
  in the testing protocol,
  where $f$ ranges over all continuous functions
  $f:\mathbf{X}\times\PPP(\mathbf{Y})\times\mathbf{Y}\to\bbbr$.
\end{theorem}

\subsection*{Proof of Theorem \ref{thm:true-asymptotic}}

We will follow the proof of Theorem \ref{thm:forecasting-asymptotic},
starting from an analogue of Lemma \ref{lem:Hilbert}.
\begin{lemma}\label{lem:true-Hilbert}
  Suppose $\mathbf{Y}$ is a metric compact.
  Let $\Phi:\mathbf{X}\times\PPP(\mathbf{Y})\times\mathbf{Y}\to\HHH$
  be a function taking values in a Hilbert space $\HHH$
  such that, for each $x$, $\Phi(x,P,y)$
  is a continuous function of $(P,y)\in\PPP(\mathbf{Y})\times\mathbf{Y}$.
  Suppose $\sup_{x,P,y}\left\|\Phi(x,P,y)\right\|_{\HHH}<\infty$
  and set
  \begin{equation*}
    \Psi
    \left(
      x,P,y
    \right)
    :=
    \Phi
    \left(
      x,P,y
    \right)
    -
    \int_{\mathbf{Y}}
    \Phi
    \left(
      x,P,y
    \right)
    P(\dd y).
  \end{equation*}
  Skeptic can force
  \begin{equation}\label{eq:true-Hilbert}
    \left\|
      \sum_{n=1}^N
      \Psi
      \left(
        x_n,P_n,y_n
      \right)
    \right\|_{\HHH}
    =
    O
    \left(
      \sqrt{N}
      \log N
    \right)
  \end{equation}
  as $N\to\infty$.
\end{lemma}
\begin{proof}
  Let
  \begin{equation*}
    c
    :=
    \sup_{x,P,y}
    \left\|
      \Psi(x,P,y)
    \right\|_{\HHH}
    <
    \infty.
  \end{equation*}
  For $k,N=1,2,\ldots$,
  define
  \begin{equation*}
    S_N^k
    :=
    \begin{cases}
      2^k+S_N & \text{if $c^2N\le2^k$}\\
      S_{N-1}^k & \text{otherwise},
    \end{cases}
  \end{equation*}
  where $S_N$ is defined as in (\ref{eq:S})
  (with $\Psi$ in place of $\Psi_n$
  in all references to the proof of Lemma \ref{lem:Hilbert}).
  Let us check that
  \begin{equation}\label{eq:series}
    S^*_N
    :=
    \sum_{k=1}^{\infty}
    k^{-2}
    2^{-k}
    S_N^k
  \end{equation}
  is a capital process (obviously non-negative) of a strategy for Skeptic
  started with a finite initial capital.
  Since $S_0^k=2^k$,
  the initial capital $\sum_{k=1}^{\infty}k^{-2}=\pi^2/6$ is indeed finite.
  It is also easy to see that the series (\ref{eq:series}) is convergent
  and that (\ref{eq:move}) still holds,
  where
  \begin{equation*}
    A
    =
    \sum_{k=K}^{\infty}
    k^{-2} 2^{-k} 2
    \sum_{n=1}^{N-1}
    \Psi(x_n,P_n,y_n)
  \end{equation*}
  for some $K$.

  Skeptic can force $S_N^*\le C$,
  where $C$ can depend on the path
  \begin{equation*}
    x_1,P_1,y_1,x_2,P_2,y_2,\ldots
  \end{equation*}
  chosen by the players
  (see Lemma 3.1 in \cite{shafer/vovk:2001} or, for a simpler argument,
  the end of the proof of Theorem 3 in \cite{\GTPVII}).
  Therefore, he can force $k^{-2}2^{-k}S_N^k\le C$ for all $k$.
  Setting $k:=\lceil \log(c^2N)\rceil$
  (with $\log$ standing for the binary logarithm),
  we can rewrite the inequality $S_N^k\le Ck^{2}2^{k}$ as
  \begin{equation*}
    2^k+S_N
    \le
    Ck^{2}2^{k},
  \end{equation*}
  which implies
  \begin{multline*}
    \left\|
      \sum_{n=1}^N
      \Psi
      \left(
        x_n,P_n,y_n
      \right)
    \right\|_{\HHH}^2
    \le
    Ck^{2}2^{k}\\
    \le
    C
    \left(
      \log(c^2N)+1
    \right)^2
    2^{\log(c^2N)+1}
    =
    O
    \left(
      N\log^2N
    \right).
    \qedmath
  \end{multline*}
\end{proof}

The following analogue of Theorem \ref{thm:forecasting-bounds}
immediately follows from Lemma \ref{lem:true-Hilbert}
and the proof of Lemma \ref{lem:RKHS}.
\begin{lemma}\label{lem:true-RKHS}
  Let $\mathbf{Y}$ be a metric compact
  and $\FFF$ be a forecast-continuous RKHS
  on $\mathbf{X}\times\PPP(\mathbf{Y})\times\mathbf{Y}$
  with finite imbedding constant.
  Skeptic can force
  \begin{equation*}
    \sum_{n=1}^N
    \left(
      f
      \left(
        x_n,P_n,y_n
      \right)
      -
      \int_{\mathbf{Y}}
      f
      \left(
        x_n,P_n,y
      \right)
      P_n(\dd y)
   \right)
    =
    O
    \left(
      \left\|
        f
      \right\|_{\FFF}
      \sqrt{N} \log N
    \right)
  \end{equation*}
  as $N\to\infty$,
  where the $O$ is uniform in $f\in\FFF$.
\end{lemma}
In its turn
Lemma \ref{lem:true-RKHS} immediately implies
the statement of Theorem \ref{thm:true-asymptotic}
in the case of compact $\mathbf{X}$ and $\mathbf{Y}$
(where the antecedent of (\ref{eq:true-calibration-cum-resolution})
is automatically true):
we can use the same argument based on (\ref{eq:approximation}).

Now let $\mathbf{X}$ and $\mathbf{Y}$ be any locally compact metric spaces.
Skeptic can use the same method based on Remover's winning strategy in the game of removal
as that used by Forecaster in the proof of Theorem \ref{thm:forecasting-asymptotic}
(see p.~\pageref{p:general}).
This completes the proof of Theorem \ref{thm:true-asymptotic}.

\subsection*{Measure-theoretic result}

In this subsection we will use some notions
of measure-theoretic probability theory,
such as regular conditional distributions;
all needed background information can be found in, e.g., \cite{shiryaev:1996}.
\begin{corollary}\label{cor:true-asymptotic}
  Suppose $\FFF_n$, $n=0,1,\ldots$, is a filtration
  (increasing sequence of $\sigma$-algebras),
  $\mathbf{X}$ and $\mathbf{Y}$ are
  compact metric spaces,
  $x_n$, $n=1,2,\ldots$, are $\FFF_{n-1}$-measurable random elements
  taking values in $\mathbf{X}$,
  $y_n$, $n=1,2,\ldots$, are $\FFF_n$-measurable random elements
  taking values in $\mathbf{Y}$,
  and $P_n\in\PPP(\mathbf{Y})$ are regular conditional distributions of $y_n$
  given $\FFF_{n-1}$.
  Then
  \begin{equation}\label{eq:true-calibration-cum-resolution-consequent}
      \forall f:
      \quad
      \lim_{N\to\infty}
      \frac1N
      \sum_{n=1}^N
      \left(
        f
        \left(
          x_n,P_n,y_n
        \right)
        -
        \int_{\mathbf{Y}}
          f
        \left(
          x_n,P_n,y
        \right)
        P_n(\dd y)
      \right)
      =
      0
  \end{equation}
  holds with probability one,
  where $f$ ranges over all continuous functions
  $f:\mathbf{X}\times\PPP(\mathbf{Y})\times\mathbf{Y}\to\bbbr$.
\end{corollary}
\begin{proof}
  Since $\mathbf{X}$ and $\mathbf{Y}$ are automatically complete and separable,
  regular conditional distributions exist
  by the corollary of Theorem II.7.5 in \cite{shiryaev:1996}.
  Our derivation of Corollary \ref{cor:true-asymptotic}
  from Theorem \ref{thm:true-asymptotic}
  will follow the standard recipe
  (\cite{shafer/vovk:2001}, Section 8.1).

  Skeptic's strategy forcing (\ref{eq:true-calibration-cum-resolution-consequent})
  (i.e., the consequent of (\ref{eq:true-calibration-cum-resolution}))
  can be chosen measurable
  (in the sense that $f_n(y)$ is a measurable function of $y$
  and the previous moves $x_1,P_1,y_1,\ldots,x_n,P_n$).
  This makes his capital process $\K_n$, $n=0,1,\ldots$,
  a martingale (in the usual measure-theoretic sense)
  with respect to the filtration $(\FFF_n)$.
  This martingale is non-negative and tends to infinity
  where (\ref{eq:true-calibration-cum-resolution-consequent}) fails;
  standard results of probability theory
  (such as Doob's inequality, \cite{shiryaev:1996}, Theorem VII.3.1.III,
  or Doob's convergence theorem, \cite{shiryaev:1996}, Theorem VII.4.1)
  imply that (\ref{eq:true-calibration-cum-resolution-consequent}) holds with probability one.
  \qedtext
\end{proof}

\section{Defensive forecasting for decision making: asymptotic theory}
\label{sec:D-idea}

Our D-prediction algorithms are built on top of probability forecasting algorithms:
D-predictions are found by minimizing the expected loss,
with the expectation taken with respect to the probability forecast.
The first problem that we have to deal with is the possibility
that the minimizer of the expected loss will be a discontinuous function,
whereas continuity is essential for the method of defensive forecasting
(cf.\ Theorem \ref{thm:forecasting-asymptotic},
where $f$ has to be a continuous function).

\subsection*{Continuity of choice functions}

It will be convenient to use the notation
\begin{equation*}
  \lambda(x,\gamma,P)
  :=
  \int_{\mathbf{Y}}
  \lambda(x,\gamma,y)
  P(\dd y),
\end{equation*}
where $P$ is a probability measure on $\mathbf{Y}$.
Let us say that $G:\mathbf{X}\times\PPP(\mathbf{Y})\to\Gamma$ is a (precise) \emph{choice function}
if it satisfies
\begin{equation*}
  \lambda(x,G(x,P),P)
  =
  \inf_{\gamma\in\Gamma}
  \lambda(x,\gamma,P),
  \quad
  \forall x\in\mathbf{X},P\in\PPP(\mathbf{Y}).
\end{equation*}
As we said,
a serious problem in implementing the expected loss minimization principle
is that there might not exist a continuous choice function $G$;
this is true even if $\mathbf{X}$, $\Gamma$, and $\mathbf{Y}$ are metric compacts
and the loss function is continuous.
If, however, the loss function $\lambda(x,\gamma,y)$ is convex in $\gamma\in\Gamma$,
there exists an approximate choice function
(although a precise choice function may still not exist).

The simplest example of a prediction game
is perhaps the \emph{simple prediction game},
in which there are no data,
$\Gamma=\mathbf{Y}=\{0,1\}$
and $\lambda(\gamma,y):=\left|y-\gamma\right|$
(omitting the $x$s from our notation).
There are no continuous approximate choice functions
in this case,
since there are no non-trivial (taking more than one value)
continuous functions from the connected space $\PPP(\mathbf{Y})$ to $\Gamma$.
If we allow randomized predictions,
the simple prediction game
effectively transforms into the following \emph{absolute loss game}:
$\Gamma=[0,1]$, $\mathbf{Y}=\{0,1\}$,
$\lambda(\gamma,y):=\left|y-\gamma\right|$.
Intuitively,
the prediction $\gamma$ in this game is the bias of the coin tossed to choose the prediction
in the simple prediction game,
and $\left|y-\gamma\right|$ is the expected loss in the latter.

Unfortunately, there is still no continuous choice function
in the absolute loss game.
It is easy to check that any choice function $G$ must satisfy
\begin{equation}\label{eq:G}
  G(P)
  :=
  \begin{cases}
    1 & \text{if $P(\{1\})>1/2$}\\
    0 & \text{if $P(\{1\})<1/2$},
  \end{cases}
\end{equation}
but the case $P(\{1\})=1/2$ is a point of bifurcation:
both predictions $\gamma=1$ and $\gamma=0$ are optimal,
as indeed is every prediction in between.
If $P(\{1\})=1/2$, the predictor finds himself in a position of Buridan's ass:
he has several equally attractive decisions to choose from.
It is clear that $G$ defined by (\ref{eq:G})
cannot be continuously extended to the whole of $\PPP(\{0,1\})$.

We have to look for approximate choice functions.
Under natural compactness and convexity conditions,
they exist by the following lemma.
\begin{lemma}\label{lem:choice}
  Let $X$ be a paracompact,
  $Y$ be a non-empty compact convex subset of a topological vector space,
  and $f:X\times Y\to\bbbr$ be a continuous function
  such that $f(x,y)$ is convex in $y\in Y$ for each $x\in X$.
  For any $\epsilon>0$ there exists a continuous ``approximate choice function'' $g:X\to Y$
  such that
  \begin{equation}\label{eq:choice}
    \forall x\in X:
    \quad
    f(x,g(x))
    \le
    \inf_{y\in Y}
    f(x,y)
    +
    \epsilon.
  \end{equation}
\end{lemma}
\begin{proof}
  Each $(x,y)\in X\times Y$ has a neighborhood $A_{x,y}\times B_{x,y}$
  such that $A_{x,y}$ and $B_{x,y}$ are open sets in $X$ and $Y$, respectively,
  and
  \begin{equation*}
    \sup_{A_{x,y}\times B_{x,y}}
    f
    -
    \inf_{A_{x,y}\times B_{x,y}}
    f
    <
    \frac{\epsilon}{2}.
  \end{equation*}
  For each $x\in X$ choose a finite subcover
  of the cover $\{A_{x,y}\times B_{x,y}\st x\in A_{x,y},y\in Y\}$ of $\{x\}\times Y$
  and let $A_x$ be the intersection of all $A_{x,y}$ in this subcover.
  The sets $A_x$ constitute an open cover of $X$ such that
  \begin{equation}\label{eq:uniform-continuity}
    \left(
      x_1\in A_x,
      x_2\in A_x
    \right)
    \Longrightarrow
    \left|
      f(x_1,y)
      -
      f(x_2,y)
    \right|
    <
    \frac{\epsilon}{2}
  \end{equation}
  for all $x\in X$ and $y\in Y$.
  Since $X$ is paracompact,
  there exists (\cite{engelking:1989}, Theorem 5.1.9)
  a locally finite partition $\{\phi_i\st i\in I\}$ of unity
  subordinated to the open cover of $X$ formed by all $A_x$, $x\in X$.
  For each $i\in I$ choose $x_i\in X$ such that $\phi_i(x_i)>0$
  (without loss of generality we can assume that such $x_i$ exists for each $i\in I$)
  and choose $y_i\in\arg\min_yf(x_i,y)$.
  Now we can set
  \begin{equation*}
    g(x)
    :=
    \sum_{i\in I}
    \phi_i(x)
    y_i.
  \end{equation*}
  Inequality (\ref{eq:choice}) follows,
  by (\ref{eq:uniform-continuity}) and the convexity of $f(x,y)$ in $y$,
  from
  \begin{multline*}
    \forall y\in Y:
    \quad
    f(x,g(x))
    =
    f
    \left(
      x,
      \sum_i
      \phi_i(x)
      y_i
    \right)
    \le
    \sum_i
    \phi_i(x)
    f
    \left(
      x,
      y_i
    \right)\\
    \le
    \sum_i
    \phi_i(x)
    f
    \left(
      x_i,
      y_i
    \right)
    +
    \frac{\epsilon}{2}
    \le
    \sum_i
    \phi_i(x)
    f
    \left(
      x_i,
      y
    \right)
    +
    \frac{\epsilon}{2}\\
    \le
    \sum_i
    \phi_i(x)
    f
    \left(
      x,
      y
    \right)
    +
    \epsilon
    =
    f(x,y)
    +
    \epsilon,
  \end{multline*}
  where $i$ ranges over the finite number of $i\in I$ for which $\phi_i(x)$ is non-zero.
  \qedtext
\end{proof}

Suppose that $\mathbf{X}$ and $\mathbf{Y}$ are compact metric spaces,
$\Gamma$ is a compact convex subset of a topological vector space,
and $\lambda(x,\gamma,y)$ is continuous in $(x,\gamma,y)$ and convex in $\gamma\in\Gamma$
(therefore, by Lemma \ref{lem:continuity2},
$\lambda(x,\gamma,P)$ is continuous in $(x,\gamma,P)\in\mathbf{X}\times\Gamma\times\PPP(\mathbf{Y})$,
and it is convex in $\gamma$).
Taking $\mathbf{X}\times\PPP(\mathbf{Y})$ as $X$ and $\Gamma$ as $Y$,
we can see that for each $\epsilon>0$
there exists an approximate choice function $G$ satisfying
\begin{equation}\label{eq:approximate-choice}
  \lambda(x,G(x,P),P)
  \le
  \inf_{\gamma\in\Gamma}
  \lambda(x,\gamma,P)
  +
  \epsilon,
  \quad
  \forall x\in\mathbf{X},P\in\PPP(\mathbf{Y}).
\end{equation}

\subsection*{Proof of a weak form of Theorem \ref{thm:decision-asymptotic}}

Suppose $\mathbf{X}$ and $\mathbf{Y}$ are compact metric spaces
and $\Gamma$ is a compact convex subset of a topological vector space.
In this subsection we will prove the existence of a prediction algorithm
guaranteeing (\ref{eq:universal-consistency})
(whose antecedent can now be ignored)
with ${}\le0$ replaced by ${}\le\epsilon$
for all continuous prediction rules $D$
for an arbitrarily small constant $\epsilon>0$.
Let $G$ satisfy (\ref{eq:approximate-choice}).
If Predictor chooses his predictions by applying the approximate choice function $G$
to $x_n$ and probability forecasts $P_n$ for $y_n$
satisfying (\ref{eq:calibration-cum-resolution}) of Theorem \ref{thm:forecasting-asymptotic},
we will have
\begin{multline}\label{eq:chain}
  \sum_{n=1}^N
  \lambda(x_n,\gamma_n,y_n)
  =
  \sum_{n=1}^N
  \lambda(x_n,G(x_n,P_n),y_n)\\
  =
  \sum_{n=1}^N
  \lambda(x_n,G(x_n,P_n),P_n)
  +
  \sum_{n=1}^N
  \Bigl(
    \lambda(x_n,G(x_n,P_n),y_n)
    -
    \lambda(x_n,G(x_n,P_n),P_n)
  \Bigr)\\
  =
  \sum_{n=1}^N
  \lambda(x_n,G(x_n,P_n),P_n)
  +
  o(N)
  \le
  \sum_{n=1}^N
  \lambda(x_n,D(x_n),P_n)
  +
  \epsilon N
  +
  o(N)\\
  =
  \sum_{n=1}^N
  \lambda(x_n,D(x_n),y_n)
  -
  \sum_{n=1}^N
  \Bigl(
    \lambda(x_n,D(x_n),y_n)
    -
    \lambda(x_n,D(x_n),P_n)
  \Bigr)
  +
  \epsilon N
  +
  o(N)\\
  =
  \sum_{n=1}^N
  \lambda(x_n,D(x_n),y_n)
  +
  \epsilon N
  +
  o(N).
\end{multline}

\section{Defensive forecasting for decision making: loss bounds}
\label{sec:D-bounds}

The goal of this section is to finish the proof of Theorem \ref{thm:decision-asymptotic}
and to establish its non-asymptotic version.
We will start with the latter.

\subsection*{Results}

Let $\FFF$ be an RKHS on $\mathbf{X}\times\mathbf{Y}$ with finite imbedding constant.
For each prediction rule $D:\mathbf{X}\to\Gamma$,
define a function $\lambda_D:\mathbf{X}\times\mathbf{Y}\to\bbbr$
by
\begin{equation*}
  \lambda_D(x,y)
  :=
  \lambda(x,D(x),y).
\end{equation*}
The notation $\left\|f\right\|_{\FFF}$ will be used
for all functions $f:\mathbf{X}\times\mathbf{Y}\to\bbbr$:
we just set $\left\|f\right\|_{\FFF}:=\infty$ for $f\notin\FFF$.
We will continue to use the notation $\ccc_{\FFF}$
for the imbedding constant
(defined by (\ref{eq:c}), where $\Omega:=\mathbf{X}\times\mathbf{Y}$).
Set
\begin{equation*}
  \ccc_{\lambda}
  :=
  \sup_{x\in\mathbf{X},\gamma\in\Gamma,y\in\mathbf{Y}}
  \lambda(x,\gamma,y)
  -
  \inf_{x\in\mathbf{X},\gamma\in\Gamma,y\in\mathbf{Y}}
  \lambda(x,\gamma,y);
\end{equation*}
this is finite if $\lambda$ is continuous
and $\mathbf{X},\Gamma,\mathbf{Y}$ are compact.
\begin{theorem}\label{thm:decision-bounds}
  Suppose $\mathbf{X}$ and $\mathbf{Y}$ are compact metric spaces,
  $\Gamma$ is a convex compact subset of a topological vector space
  and the loss function $\lambda(x,\gamma,y)$ is continuous in $(x,\gamma,y)$
  and convex in $\gamma\in\Gamma$.
  Let $\FFF$ be a forecast-continuous RKHS on $\mathbf{X}\times\mathbf{Y}$
  with finite imbedding constant $\ccc_{\FFF}$.
  There is an on-line prediction algorithm that guarantees
  \begin{equation}\label{eq:decision-bounds}
    \sum_{n=1}^N
    \lambda(x_n,\gamma_n,y_n)
    \le
    \sum_{n=1}^N
    \lambda(x_n,D(x_n),y_n)
    +
    \sqrt
    {
      \ccc_{\lambda}^2
      +
      4\ccc_{\FFF}^2
    }
    \left(
      \left\|
        \lambda_D
      \right\|_{\FFF}
      +
      1
    \right)
    \sqrt{N}
    +
    1
  \end{equation}
  for all prediction rules $D$ and all $N=1,2,\ldots$\,.
\end{theorem}

An application of Hoeffding's inequality immediately gives
the following corollary
(we postpone the details of the simple proof until p.~\pageref{p:proof}).
\begin{corollary}\label{cor:decision-bounds}
  Suppose $\mathbf{X},\Gamma,\mathbf{Y}$ are compact metric spaces
  and the loss function $\lambda$ is continuous.
  Let $N\in\{1,2,\ldots\}$ and $\delta\in(0,1)$.
  There is a randomized on-line prediction algorithm achieving
  \begin{multline*}
    \sum_{n=1}^N
    \lambda(x_n,g_n,y_n)
    \le
    \sum_{n=1}^N
    \lambda(x_n,d_n,y_n)\\
    +
    \sqrt
    {
      \ccc_{\lambda}^2
      +
      4\ccc_{\FFF}^2
    }
    \left(
      \left\|
        \lambda_D
      \right\|_{\FFF}
      +
      1
    \right)
    \sqrt{N}
    +
    \ccc_{\lambda}
    \sqrt{2\ln\frac{1}{\delta}}
    \sqrt{N}
    +
    1
  \end{multline*}
  with probability at least $1-\delta$
  for any randomized prediction rule $D:\mathbf{X}\to\PPP(\Gamma)$;
  $g_n$ and $d_n$ are independent random variables
  distributed as $\gamma_n$ and $D(x_n)$, respectively.
\end{corollary}

The above results are non-vacuous only when $\lambda_D$ is an element of the function space $\FFF$.
If $\FFF$ is a Sobolev space,
this condition follows from $D$ being in the Sobolev space
and the smoothness of $\lambda$.
For example,
Moser proved in 1966 the following result concerning composition in Sobolev spaces.
Let $\Omega$ be a smooth bounded domain in $\bbbr^K$
and $m$ be an integer number satisfying $2m>K$.
If $u\in H^{m}(\Omega)$ and $\Phi\in C^{m}(\bbbr)$,
then $\Phi \circ u\in H^{m}(\Omega)$
(see \cite{moser:1966};
for further results, see \cite{brezis/mironescu:2001}).

\subsection*{Two special cases of calibration-cum-resolution}

In the chain (\ref{eq:chain}) we applied the law of large numbers
(the property of good calibration-cum-resolution) twice:
in the third and fifth equalities.
It is easy to see, however, that in fact
the fifth equality depends only on resolution
and the third equality, although it depends on calibration-cum-resolution,
involves a known function $f$
(in the notation of (\ref{eq:calibration-cum-resolution})).
We will say that the fifth equality depends on ``general resolution''
whereas the third equality depends on ``specific calibration-cum-resolution''.
This limited character of the required calibration-cum-resolution
becomes important for obtaining good bounds on the predictive performance:
in the following subsections we will construct prediction algorithms
that satisfy the properties of specific calibration-cum-resolution and general resolution
and merge them into one algorithm;
we will start from the last step.

\subsection*{Synthesis of prediction algorithms}

The following corollary of Lemma \ref{lem:Hilbert}
will allow us to construct prediction algorithms
that achieve two goals simultaneously
(specific calibration-cum-resolution and general resolution).
\begin{corollary}\label{cor:mixture}
  Let $\mathbf{Y}$ be a metric compact
  and $\Phi_{n,j}:\mathbf{X}\times\PPP(\mathbf{Y})\times\mathbf{Y}\to\HHH_j$,
  $n=1,2,\ldots$, $j=0,1$,
  be functions taking values in Hilbert spaces $\HHH_j$
  and such that $\Phi_{n,j}(x,P,y)$ is continuous in $(P,y)$ for all $n$ and both $j$.
  Let $a_0$ and $a_1$ be two positive constants.
  There is a probability forecasting strategy that guarantees
  \begin{multline*}
    \left\|
      \sum_{n=1}^N
      \Psi_{n,j}(x_n,P_n,y_n)
    \right\|^2_{\HHH_j}\\
    \le
    \frac{1}{a_j}
    \sum_{n=1}^N
    \left(
      a_0
      \left\|
        \Psi_{n,0}(x_n,P_n,y_n)
      \right\|^2_{\HHH_0}
      +
      a_1
      \left\|
        \Psi_{n,1}(x_n,P_n,y_n)
      \right\|^2_{\HHH_1}
    \right)
  \end{multline*}
  for all $N$ and for both $j=0$ and $j=1$,
  where
  \begin{equation*}
    \Psi_{n,j}
    \left(
      x,P,y
    \right)
    :=
    \Phi_{n,j}
    \left(
      x,P,y
    \right)
    -
    \int_{\mathbf{Y}}
    \Phi_{n,j}
    \left(
      x,P,y
    \right)
    P(\dd y).
  \end{equation*}
\end{corollary}
\begin{proof}
  Define the ``weighted direct sum'' $\HHH$ of $\HHH_0$ and $\HHH_1$
  as the Cartesian product $\HHH_0\times\HHH_1$
  equipped with the inner product
  \begin{equation*}
    \langle g, g' \rangle_{\HHH}
    =
    \left\langle
      (g_0,g_1),
      (g'_0,g'_1)
    \right\rangle_{\HHH}
    :=
    \sum_{j=0}^{1}
    a_j
    \langle g_j, g'_j\rangle_{\HHH_j}.
  \end{equation*}
  Now we can define $\Phi:\mathbf{X}\times\PPP(\mathbf{Y})\times\mathbf{Y}\to\HHH$ by
  \begin{equation*}
    \Phi_n(x,P,y)
    :=
    \left(
      \Phi_{n,0}(x,P,y),
      \Phi_{n,1}(x,P,y)
    \right).
  \end{equation*}
  It is clear that $\Phi_{n}(x,P,y)$ is continuous in $(P,y)$ for all $n$.
  Applying the strategy of Lemma \ref{lem:Hilbert} to it
  and using~(\ref{eq:Hilbert}),
  we obtain
  \begin{multline*}
    a_j
    \left\|
      \sum_{n=1}^N
      \Psi_{n,j}(x_n,P_n,y_n)
    \right\|^2_{\HHH_j}\\
    \le
    \left\|
      \left(
        \sum_{n=1}^N
        \Psi_{n,0}(x_n,P_n,y_n),
        \sum_{n=1}^N
        \Psi_{n,1}(x_n,P_n,y_n)
      \right)
    \right\|^2_{\HHH}\\
    =
    \left\|
      \sum_{n=1}^N
      \Psi_n(x_n,P_n,y_n)
    \right\|^2_{\HHH}
    \le
    \sum_{n=1}^N
    \left\|
      \Psi_n(x_n,P_n,y_n)
    \right\|^2_{\HHH}\\
    =
    \sum_{n=1}^N
    \sum_{j=0}^{1}
    a_j
    \left\|
      \Psi_{n,j}(x_n,P_n,y_n)
    \right\|^2_{\HHH_j}.
    \qedmath
  \end{multline*}
\end{proof}

Suppose $\mathbf{X},\Gamma,\mathbf{Y}$ are metric compacts
and $\FFF$ is a forecast-continuous RKHS on $\mathbf{X}\times\mathbf{Y}$.
Let $G_n:\mathbf{X}\times\PPP(\mathbf{Y})\to\Gamma$
be a sequence of approximate choice functions
satisfying
\begin{equation*}
  \lambda(x,G_n(x,P),P)
  <
  \inf_{\gamma\in\Gamma}
  \lambda(x,\gamma,P)
  +
  2^{-n},
  \quad
  \forall x\in\mathbf{X},P\in\PPP(\mathbf{Y})
\end{equation*}
(they exist by (\ref{eq:approximate-choice})).
Corollary \ref{cor:mixture} will be applied to $a_0=a_1=1$ and to the mappings
\begin{align}
  \Psi_{n,0}(x,P,y)
  &:=
  \lambda(x,G_n(x,P),y)
  -
  \lambda(x,G_n(x,P),P),
  \label{eq:Psi-0}\\
  \Psi_{n,1}(x,P,y)
  &:=
  \kkk_{x,y}-\kkk_{x,P},
  \label{eq:Psi-1}
\end{align}
where $\kkk_{x,y}$ is the evaluation functional at $(x,y)$ for $\FFF$
and $\kkk_{x,P}$ is the mean of $\kkk_{x,y}$ with respect to $P(\dd y)$.
It is easy to see that
\begin{equation}\label{eq:constants}
  \left\|
    \Psi_{n,0}(x,P,y)
  \right\|_{\bbbr}
  =
  \left|
    \Psi_{n,0}(x,P,y)
  \right|
  \le
  \ccc_{\lambda},
  \quad
  \left\|
    \Psi_{n,1}(x,P,y)
  \right\|_{\FFF}
  \le
  2\ccc_{\FFF}.
\end{equation}

\subsection*{Specific calibration-cum-resolution}

Corollary \ref{cor:mixture} immediately implies:
\begin{lemma}
  The probability forecasting strategy of Corollary \ref{cor:mixture}
  based on (\ref{eq:Psi-0}) and (\ref{eq:Psi-1}) guarantees
  \begin{equation*}
    \left|
      \sum_{n=1}^N
      \Bigl(
        \lambda(x_n,G_n(x_n,P_n),y_n)
        -
        \lambda(x_n,G_n(x_n,P_n),P_n)
      \Bigr)
    \right|
    \le
    \sqrt{\ccc_{\lambda}^2+4\ccc_{\FFF}^2}
    \sqrt{N}.
  \end{equation*}
\end{lemma}
\begin{proof}
  This follows from
  \begin{equation*}
    \left|
      \sum_{n=1}^N
      \Bigl(
        \lambda(x_n,G_n(x_n,P_n),y_n)
        -
        \lambda(x_n,G_n(x_n,P_n),P_n)
      \Bigr)
    \right|^2
    \le
    \sum_{n=1}^N
    \left(
      \ccc_{\lambda}^2
      +
      4\ccc_{\FFF}^2
    \right)
  \end{equation*}
  (see (\ref{eq:constants})).
  \qedtext
\end{proof}

\subsection*{General resolution I}

The following lemma is proven similarly to Lemma \ref{lem:RKHS}.
\begin{lemma}\label{lem:resolution-1}
  The probability forecasting strategy of Corollary \ref{cor:mixture}
  based on (\ref{eq:Psi-0}) and (\ref{eq:Psi-1}) guarantees
  \begin{equation*}
    \left|
      \sum_{n=1}^N
      \Bigl(
        \lambda(x_n,D(x_n),y_n)
        -
        \lambda(x_n,D(x_n),P_n)
      \Bigr)
    \right|
    \le
    \sqrt{\ccc_{\lambda}^2+4\ccc_{\FFF}^2}
    \left\|
      \lambda_D
    \right\|_{\FFF}
    \sqrt{N}.
  \end{equation*}
\end{lemma}
\begin{proof}
  This follows from
  \begin{multline*}
    \left|
      \sum_{n=1}^N
      \Bigl(
        \lambda(x_n,D(x_n),y_n)
        -
        \lambda(x_n,D(x_n),P_n)
      \Bigr)
    \right|\\
    =
    \left|
      \sum_{n=1}^N
      \Bigl(
        \lambda_D(x_n,y_n)
        -
        \lambda_D(x_n,P_n)
      \Bigr)
    \right|\\
    =
    \left|
      \sum_{n=1}^N
      \left\langle
        \lambda_D,
        \kkk_{x_n,y_n}-\kkk_{x_n,P_n}
      \right\rangle_{\FFF}
    \right|
    \le
    \left\|
      \lambda_D
    \right\|_{\FFF}
    \left\|
      \sum_{n=1}^N
      \left(
        \kkk_{x_n,y_n}-\kkk_{x_n,P_n}
      \right)
    \right\|_{\FFF}\\
    \le
    \left\|
      \lambda_D
    \right\|_{\FFF}
    \sqrt
    {
      \sum_{n=1}^N
      \left(
        \ccc_{\lambda}^2+4\ccc_{\FFF}^2
      \right)
    }
    =
    \sqrt
    {
      \ccc_{\lambda}^2
      +
      4\ccc_{\FFF}^2
    }
    \left\|
      \lambda_D
    \right\|_{\FFF}
    \sqrt{N}
  \end{multline*}
  (we have used Corollary \ref{cor:mixture} and (\ref{eq:constants})).
  \qedtext
\end{proof}

\subsection*{Proof of Theorem \ref{thm:decision-bounds}}

Let $\gamma_n:=G_n(x_n,P_n)$
where $P_n$ are produced by the probability forecasting strategy of Corollary \ref{cor:mixture}
based on (\ref{eq:Psi-0}) and (\ref{eq:Psi-1}).
Following (\ref{eq:chain})
and using the previous two lemmas, we obtain:
\begin{align*}
  \sum_{n=1}^N
  \lambda(x_n,\gamma_n,y_n)
  &=
  \sum_{n=1}^N
  \lambda(x_n,G_n(x_n,P_n),y_n)\\
  &=
  \sum_{n=1}^N
  \lambda(x_n,G_n(x_n,P_n),P_n)\\
  &\quad{}+
  \sum_{n=1}^N
  \Bigl(
    \lambda(x_n,G_n(x_n,P_n),y_n)
    -
    \lambda(x_n,G_n(x_n,P_n),P_n)
  \Bigr)\\
  &\le
  \sum_{n=1}^N
  \lambda(x_n,G_n(x_n,P_n),P_n)
  +
  \sqrt
  {
    \ccc_{\lambda}^2
    +
    4\ccc_{\FFF}^2
  }
  \sqrt{N}\\
  &\le
  \sum_{n=1}^N
  \lambda(x_n,D(x_n),P_n)
  +
  \sqrt
  {
    \ccc_{\lambda}^2
    +
    4\ccc_{\FFF}^2
  }
  \sqrt{N}
  +
  1\\
  &=
  \sum_{n=1}^N
  \lambda(x_n,D(x_n),y_n)
  +
  \sqrt
  {
    \ccc_{\lambda}^2
    +
    4\ccc_{\FFF}^2
  }
  \sqrt{N}
  +
  1\\
  &\quad{}-
  \sum_{n=1}^N
  \Bigl(
    \lambda(x_n,D(x_n),y_n)
    -
    \lambda(x_n,D(x_n),P_n)
  \Bigr)\\
  &\le
  \sum_{n=1}^N
  \lambda(x_n,D(x_n),y_n)
  +
  \sqrt
  {
    \ccc_{\lambda}^2
    +
    4\ccc_{\FFF}^2
  }
  \left(
    \left\|
      \lambda_D
    \right\|_{\FFF}
    +
    1
  \right)
  \sqrt{N}
  +
  1.
\end{align*}

\subsection*{Proof of Corollary \ref{cor:decision-bounds}}

\label{p:proof}Since $\lambda(x_n,g_n,y_n)-\lambda(x_n,d_n,y_n)$
never exceeds $\ccc_{\lambda}$ in absolute value,
Hoeffding's inequality (\cite{cesabianchi/lugosi:2006}, Corollary A.1)
shows that
\begin{multline*}
  \Prob
  \Biggl\{
    \sum_{n=1}^N
    \Bigl(
      \lambda(x_n,g_n,y_n)
      -
      \lambda(x_n,d_n,y_n)
    \Bigr)
    -
    \sum_{n=1}^N
    \Bigl(
      \lambda(x_n,\gamma_n,y_n)
      -
      \lambda(x_n,D(x_n),y_n)
    \Bigr)\\
    >
    t
  \Biggr\}
  \le
  \exp
  \left(
    -\frac{t^2}{2\ccc_{\lambda}^2N}
  \right)
\end{multline*}
for every $t>0$.
Choosing $t$ satisfying
\begin{equation*}
  \exp
  \left(
    -\frac{t^2}{2\ccc_{\lambda}^2N}
  \right)
  =
  \delta,
\end{equation*}
i.e.,
\begin{equation*}
  t
  :=
  \ccc_{\lambda}
  \sqrt{2\ln\frac{1}{\delta}}
  \sqrt{N},
\end{equation*}
we obtain the statement of Corollary \ref{cor:decision-bounds}.

\subsection*{General resolution II}

To prove Theorem \ref{thm:decision-asymptotic},
we will need the following variation on Lemma \ref{lem:resolution-1}.
\begin{lemma}\label{lem:resolution-2}
  The probability forecasting strategy of Corollary \ref{cor:mixture}
  based on (\ref{eq:Psi-0}) and (\ref{eq:Psi-1}) guarantees
  \begin{equation*}
    \left|
      \sum_{n=1}^N
      \left(
        f(x_n,y_n)
        -
        \int_{\mathbf{Y}}
          f(x_n,y)
        P(\dd y)
      \right)
    \right|
    \le
    \sqrt{\ccc_{\lambda}^2+4\ccc_{\FFF}^2}
    \left\|
      f
    \right\|_{\FFF}
    \sqrt{N}
  \end{equation*}
  for any $f\in\FFF$.
\end{lemma}
\begin{proof}
  Following the proof of Lemma \ref{lem:resolution-1}:
  \begin{multline*}
    \left|
      \sum_{n=1}^N
      \left(
        f(x_n,y_n)
        -
        \int_{\mathbf{Y}}
          f(x_n,y)
        P_n(\dd y)
      \right)
    \right|\\
    =
    \left|
      \sum_{n=1}^N
      \left\langle
        f,
        \kkk_{x_n,y_n}-\kkk_{x_n,P_n}
      \right\rangle_{\FFF}
    \right|
    \le
    \left\|
      f
    \right\|_{\FFF}
    \left\|
      \sum_{n=1}^N
      \left(
        \kkk_{x_n,y_n}-\kkk_{x_n,P_n}
      \right)
    \right\|_{\FFF}\\
    \le
    \left\|
      f
    \right\|_{\FFF}
    \sqrt
    {
      \sum_{n=1}^N
      \left(
        \ccc_{\lambda}^2+4\ccc_{\FFF}^2
      \right)
    }
    =
    \sqrt
    {
      \ccc_{\lambda}^2
      +
      4\ccc_{\FFF}^2
    }
    \left\|
      f
    \right\|_{\FFF}
    \sqrt{N}.
    \qedmath
  \end{multline*}
\end{proof}

\subsection*{Proof of Theorem \ref{thm:decision-asymptotic}}

As in the proof of Theorem \ref{thm:forecasting-asymptotic},
we first assume that $\mathbf{X}$, $\Gamma$, and $\mathbf{Y}$ are compact.
Let us first see that the prediction algorithm of Theorem \ref{thm:decision-bounds}
fed with a suitable RKHS
guarantees the consequent of (\ref{eq:universal-consistency})
for all continuous $D$.
Let $\FFF$ be a universal and continuous RKHS on $\mathbf{X}\times\mathbf{Y}$
with finite imbedding constant $\ccc_{\FFF}$.

Fix a continuous decision rule $D:\mathbf{X}\to\Gamma$.
For any $\epsilon>0$,
we can find a function $f\in\FFF$
that is $\epsilon$-close in $C(\mathbf{X}\times\mathbf{Y})$ to $\lambda(x,D(x),y)$.
Following (\ref{eq:chain}) and the similar chain
in the proof of Theorem \ref{thm:decision-bounds},
we obtain:
\begin{align*}
  \sum_{n=1}^N
  \lambda(x_n,\gamma_n,y_n)
  &=
  \sum_{n=1}^N
  \lambda(x_n,G_n(x_n,P_n),y_n)\\
  &=
  \sum_{n=1}^N
  \lambda(x_n,G_n(x_n,P_n),P_n)\\
  &\quad{}+
  \sum_{n=1}^N
  \Bigl(
    \lambda(x_n,G_n(x_n,P_n),y_n)
    -
    \lambda(x_n,G_n(x_n,P_n),P_n)
  \Bigr)\\
  &\le
  \sum_{n=1}^N
  \lambda(x_n,G_n(x_n,P_n),P_n)
  +
  \sqrt
  {
    \ccc_{\lambda}^2
    +
    4\ccc_{\FFF}^2
  }
  \sqrt{N}\\
  &\le
  \sum_{n=1}^N
  \lambda(x_n,D(x_n),P_n)
  +
  \sqrt
  {
    \ccc_{\lambda}^2
    +
    4\ccc_{\FFF}^2
  }
  \sqrt{N}
  +
  1\\
  &=
  \sum_{n=1}^N
  \lambda(x_n,D(x_n),y_n)
  +
  \sqrt
  {
    \ccc_{\lambda}^2
    +
    4\ccc_{\FFF}^2
  }
  \sqrt{N}
  +
  1\\
  &\quad{}-
  \sum_{n=1}^N
  \Bigl(
    \lambda(x_n,D(x_n),y_n)
    -
    \lambda(x_n,D(x_n),P_n)
  \Bigr)\\
  &\le
  \sum_{n=1}^N
  \lambda(x_n,D(x_n),y_n)
  +
  \sqrt
  {
    \ccc_{\lambda}^2
    +
    4\ccc_{\FFF}^2
  }
  \sqrt{N}
  +
  1\\
  &\quad{}-
  \sum_{n=1}^N
  \Bigl(
    f(x_n,y_n)
    -
    \int_{\mathbf{Y}}
      f(x_n,y)
    P_n(y)
  \Bigr)
  +
  2\epsilon N\\
  &\le
  \sum_{n=1}^N
  \lambda(x_n,D(x_n),y_n)
  +
  \sqrt
  {
    \ccc_{\lambda}^2
    +
    4\ccc_{\FFF}^2
  }
  \left(
    \left\|
      f
    \right\|_{\FFF}
    +
    1
  \right)
  \sqrt{N}
  +
  1\\
  &\quad{}+
  2\epsilon N.
\end{align*}
We can see that
\begin{equation*}
  \limsup_{N\to\infty}
  \left(
    \frac1N
    \sum_{n=1}^N
    \lambda(x_n,\gamma_n,y_n)
    -
    \frac1N
    \sum_{n=1}^N
    \lambda(x_n,D(x_n),y_n)
  \right)
  \le
  2\epsilon;
\end{equation*}
since this is true for any $\epsilon>0$,
the consequent of (\ref{eq:universal-consistency}) holds.

It remains to get rid of the assumption of compactness
of $\mathbf{X}$, $\Gamma$, and $\mathbf{Y}$.
We will need the following lemma.
\begin{lemma}\label{lem:C-det}
  Under the conditions of Theorem \ref{thm:decision-asymptotic},
  for each pair of compact sets $A\subseteq\mathbf{X}$ and $B\subseteq\mathbf{Y}$
  there exists a compact set $C=C(A,B)\subseteq\Gamma$
  such that for each continuous prediction rule $D:\mathbf{X}\to\Gamma$
  there exists a continuous prediction rule $D':\mathbf{X}\to C$
  that dominates $D$ in the sense
  \begin{equation}\label{eq:prediction-type}
    \forall x\in A,y\in B:
    \quad
    \lambda(x,D'(x),y)
    \le
    \lambda(x,D(x),y).
  \end{equation}
\end{lemma}
\begin{proof}
  Without loss of generality $A$ and $B$ are assumed non-empty.
  Fix any $\gamma_0\in\Gamma$.
  Let
  \begin{equation*}
    M_1
    :=
    \sup_{(x,y)\in A\times B}
    \lambda(x,\gamma_0,y),
  \end{equation*}
  let $C_1\subseteq\Gamma$ be a compact set such that  
  \begin{equation*}
    \forall x\in A,\gamma\notin C_1,y\in B:
    \quad
    \lambda(x,\gamma,y)
    >
    M_1+1,
  \end{equation*}
  let
  \begin{equation*}
    M_2
    :=
    \sup_{(x,\gamma,y)\in A\times C_1\times B}
    \lambda(x,\gamma,y).
  \end{equation*}
  and let $C_2\subseteq\Gamma$ be a compact set such that  
  \begin{equation*}
    \forall x\in A,\gamma\notin C_2,y\in B:
    \quad
    \lambda(x,\gamma,y)
    >
    M_2+1.
  \end{equation*}
  It is obvious that $M_1\le M_2$ and $\gamma_0\in C_1\subseteq C_2$.

  Let us now check that $C_1$ lies inside the interior of $C_2$.
  Indeed, for any fixed $(x,y)\in A\times B$ and $\gamma\in C_1$,
  we have $\lambda(x,\gamma,y)\le M_2$;
  since $\lambda(x,\gamma',y)>M_2+1$ for all $\gamma'\notin C_2$,
  some neighborhood of $\gamma$ will lie completely in $C_2$.

  Let $D:\mathbf{X}\to\Gamma$ be a continuous prediction rule.
  We will show that (\ref{eq:prediction-type}) holds for some continuous prediction rule $D'$
  taking values in the compact set $C_2$.
  Namely,
  we define
  \begin{multline*}
    D'(x)
    :=\\
    \begin{cases}
      D(x) & \text{if $D(x)\in C_1$}\\
      \frac{\rho(D(x),\Gamma\setminus C_2)}{\rho(D(x),C_1)+\rho(D(x),\Gamma\setminus C_2)} D(x)
      +\frac{\rho(D(x),C_1)}{\rho(D(x),C_1)+\rho(D(x),\Gamma\setminus C_2)} \gamma_0
      & \text{if $D(x)\in C_2\setminus C_1$}\\
      \gamma_0 & \text{if $D(x)\in \Gamma\setminus C_2$}
    \end{cases}
  \end{multline*}
  where $\rho$ is the metric on $\Gamma$;
  the denominator $\rho(D(x),C_1)+\rho(D(x),\Gamma\setminus C_2)$
  is always positive since already $\rho(D(x),C_1)$ is positive.
  Assuming $C_2$ convex
  (which can be done by \cite{rudin:1991}, Theorem 3.20(c)),
  we can see that $D'$ indeed takes values in $C_2$.
  The only points $x$ at which the continuity of $D'$ is not obvious
  are those for which $D(x)$ lies on the boundary of $C_1$:
  one has to use the fact that $C_1$ is covered by the interior of $C_2$.

  It remains to check (\ref{eq:prediction-type});
  the only non-trivial case is $D(x)\in C_2\setminus C_1$.
  By the convexity of $\lambda(x,\gamma,y)$ in $\gamma$,
  the inequality in (\ref{eq:prediction-type}) will follow from
  \begin{multline*}
    \frac{\rho(D(x),\Gamma\setminus C_2)}{\rho(D(x),C_1)+\rho(D(x),\Gamma\setminus C_2)}
    \lambda(x,D(x),y)\\
    +\frac{\rho(D(x),C_1)}{\rho(D(x),C_1)+\rho(D(x),\Gamma\setminus C_2)}
    \lambda(x,\gamma_0,y)
    \le
    \lambda(x,D(x),y),
  \end{multline*}
  i.e.,
  \begin{equation*}
    \lambda(x,\gamma_0,y)
    \le
    \lambda(x,D(x),y).
  \end{equation*}
  Since the left-hand side of the last inequality is at most $M_1$
  and its right-hand side exceeds $M_1+1$,
  it holds true.
  \qedtext
\end{proof}
For each pair of compact $A\subseteq\mathbf{X}$ and $B\subseteq\mathbf{Y}$
fix a compact $C(A,B)\subseteq\Gamma$ as in the lemma.
Similarly to the proof of Theorem \ref{thm:forecasting-asymptotic},
Predictor's strategy ensuring (\ref{eq:universal-consistency})
is constructed from Remover's winning strategy in $G(\mathbf{X}\times\mathbf{Y})$
and from Predictor's strategies $\SSS(A,B)$ outputting predictions $\gamma_n\in C(A,B)$
and ensuring the consequent of (\ref{eq:universal-consistency})
for $D:A\to C(A,B)$
under the assumption that $(x_n,y_n)\in A\times B$
for given compact $A\subseteq\mathbf{X}$ and $B\subseteq\mathbf{Y}$.
Remover's moves are assumed to be of the form $A\times B$
for compact $A\subseteq\mathbf{X}$ and $B\subseteq\mathbf{Y}$.
Predictor is simultaneously playing the game of removal
$G(\mathbf{X}\times\mathbf{Y})$ as Evader.

Predictor asks Remover to make his first move $A_1\times B_1$ in the game of removal.
Predictor then plays the prediction game using the strategy $\SSS(A_1,B_1)$
until Reality chooses $(x_n,y_n)\notin A_1\times B_1$
(forever if Reality never chooses such $(x_n,y_n)$).
As soon as such $(x_n,y_n)$ is chosen,
Predictor announces $(x_n,y_n)$ in the game of removal
and notes Remover's response $(A_2,B_2)$.
He then continues playing the prediction game using the strategy $\SSS(A_2,B_2)$
until Reality chooses $(x_n,y_n)\notin A_2\times B_2$,
etc.

Let us check that this strategy for Predictor
will always ensure (\ref{eq:universal-consistency}).
If Reality chooses $(x_n,y_n)$ outside Predictor's current $A_k\times B_k$
finitely often,
the consequent of (\ref{eq:universal-consistency}) will be satisfied
for all continuous $D:\mathbf{X}\to C(A_K,B_K)$
($(A_K,B_K)$ being Remover's last move)
and so, by Lemma \ref{lem:C-det},
for all continuous $D:\mathbf{X}\to\Gamma$.
If Reality chooses $(x_n,y_n)$ outside Predictor's current $A_k\times B_k$
infinitely often,
the set of $(x_n,y_n)$, $n=1,2,\ldots$, will not be precompact,
and so the antecedent of (\ref{eq:universal-consistency}) will be violated.

\subsection*{Proof of Theorem \ref{cor:decision-asymptotic}}

Define
\begin{equation}\label{eq:expected-loss}
  \lambda(x,\gamma,y)
  :=
  \int_{\Gamma}
  \lambda(x,g,y)
  \gamma(\dd g),
\end{equation}
where $\gamma$ is a probability measure on $\Gamma$.
This is the loss function in a new game of prediction
with the prediction space $\PPP(\Gamma)$.
When $\gamma$ ranges over $\PPP(C)$
(identified with the subset of $\PPP(\Gamma)$
consisting of the measures concentrated on $C$)
for a compact $C$,
the loss function (\ref{eq:expected-loss}) is continuous
by Lemma \ref{lem:continuity2}.
We need the following analogue of Lemma \ref{lem:C-det}.
\begin{lemma}\label{lem:C-rand}
  Under the conditions of Theorem \ref{cor:decision-asymptotic},
  for each pair of compact sets $A\subseteq\mathbf{X}$ and $B\subseteq\mathbf{Y}$
  there exists a compact set $C=C(A,B)\subseteq\Gamma$
  such that for each continuous randomized prediction rule $D:\mathbf{X}\to\PPP(\Gamma)$
  there exists a continuous randomized prediction rule $D':\mathbf{X}\to\PPP(C)$
  such that (\ref{eq:prediction-type}) holds
  ($D'$ dominates $D$ ``on average'').
\end{lemma}
\begin{proof}
  Define $\gamma_0$, $C_1$, and $C_2$ as in the proof of Lemma \ref{lem:C-det}.
  Fix a continuous function $f_1:\Gamma\to[0,1]$ such that $f_1=1$ on $C_1$
  and $f_1=0$ on $\Gamma\setminus C_2$
  (such an $f_1$ exists by the Tietze--Uryson theorem,
  \cite{engelking:1989}, 2.1.8).
  Set $f_2:=1-f_1$.
  Let $D:\mathbf{X}\to\PPP(\Gamma)$ be a continuous randomized prediction rule.
  For each $x\in\mathbf{X}$,
  split $D(x)$ into two measures on $\Gamma$
  absolutely continuous with respect to $D(x)$:
  $D_1(x)$ with Radon--Nikodym density $f_1$
  and $D_2(x)$ with Radon--Nikodym density $f_2$;
  set
  \begin{equation*}
    D'(x)
    :=
    D_1(x)
    +
    \left|D_2(x)\right|
    \delta_{\gamma_0}
  \end{equation*}
  (letting $\left|P\right|:=P(\Gamma)$ for $P\in\PPP(\Gamma)$).
  It is clear that $D'$ is continuous
  (in the topology of weak convergence, as usual),
  takes values in $\PPP(C_2)$,
  and
  \begin{multline*}
    \lambda(x,D'(x),y)
    =
    \int_{\Gamma}
      \lambda(x,\gamma,y)
      f_1(\gamma)
    D(x)(\dd\gamma)
    +
    \lambda(x,\gamma_0,y)
    \int_{\Gamma}
      f_2(\gamma)
    D(x)(\dd\gamma)\\
    \le
    \int_{\Gamma}
      \lambda(x,\gamma,y)
      f_1(\gamma)
    D(x)(\dd\gamma)
    +
    \int_{\Gamma}
      M_1
      f_2(\gamma)
    D(x)(\dd\gamma)\\
    \le
    \int_{\Gamma}
      \lambda(x,\gamma,y)
      f_1(\gamma)
    D(x)(\dd\gamma)
    +
    \int_{\Gamma}
      \lambda(x,\gamma,y)
      f_2(\gamma)
    D(x)(\dd\gamma)
    =
    \lambda(x,D(x),y)
  \end{multline*}
  for all $(x,y)\in A\times B$.
  \qedtext
\end{proof}
Fix one of the mappings $(A,B)\mapsto C(A,B)$
whose existence is asserted by the lemma.

We will prove that the strategy of the previous subsection
with $\PPP(C(A,B))$ in place of $C(A,B)$
applied to the new game
is universally consistent.
Let $D:\mathbf{X}\to\PPP(\Gamma)$ be a continuous randomized prediction rule,
i.e., a continuous prediction rule in the new game.
Let $(A_K,B_K)$ be Remover's last move
(if Remover makes infinitely many moves,
the antecedent of (\ref{eq:universal-consistency-randomized}) is false,
and there is nothing to prove),
and let $D':\mathbf{X}\to\PPP(C(A_K,B_K))$ be a continuous randomized prediction rule
satisfying (\ref{eq:prediction-type}) with $A:=A_K$ and $B:=B_K$.
From some $n$ on
our randomized prediction algorithm produces $\gamma_n\in\PPP(\Gamma)$
concentrated on $C(A_K,B_K)$,
and they will satisfy
\begin{multline}\label{eq:stage-K}
  \limsup_{N\to\infty}
  \left(
    \frac1N
    \sum_{n=1}^N
    \lambda(x_n,\gamma_n,y_n)
    -
    \frac1N
    \sum_{n=1}^N
    \lambda(x_n,D(x_n),y_n)
  \right)\\
  \le
  \limsup_{N\to\infty}
  \left(
    \frac1N
    \sum_{n=1}^N
    \lambda(x_n,\gamma_n,y_n)
    -
    \frac1N
    \sum_{n=1}^N
    \lambda(x_n,D'(x_n),y_n)
  \right)
  \le
  0.
\end{multline}

The loss function is bounded in absolute value
on the compact set
$A_K\times\left( C(A_K,B_K)\cup D(A_K)\right)\times B_K$
by a constant $c$.
The law of the iterated logarithm
(see, e.g., \cite{shafer/vovk:2001}, (5.8))
implies that
\begin{align*}
  \limsup_{N\to\infty}
  \frac
  {
    \left|
      \sum_{n=1}^N
      \left(
        \lambda(x_n,g_n,y_n)
        -
        \lambda(x_n,\gamma_n,y_n)
      \right)
    \right|
  }
  {
    \sqrt{2c^2N\ln\ln N}
  }
  &\le
  1,\\
  \limsup_{N\to\infty}
  \frac
  {
    \left|
      \sum_{n=1}^N
      \left(
        \lambda(x_n,d_n,y_n)
        -
        \lambda(x_n,D(x_n),y_n)
      \right)
    \right|
  }
  {
    \sqrt{2c^2N\ln\ln N}
  }
  &\le
  1
\end{align*}
with probability one.
Combining the last two inequalities with (\ref{eq:stage-K}) gives
\begin{equation*}
  \limsup_{N\to\infty}
  \left(
    \frac1N
    \sum_{n=1}^N
    \lambda(x_n,g_n,y_n)
    -
    \frac1N
    \sum_{n=1}^N
    \lambda(x_n,d_n,y_n)
  \right)
  \le
  0
  \enspace
  \textrm{a.s.}
\end{equation*}
This immediately implies (\ref{eq:universal-consistency-randomized}).

\section{Conclusion}
\label{sec:conclusion}

In this section I will list
what I think are interesting directions of further research.

\subsection*{The data space as a bottleneck}

It is easy to see that if we set
$\mathbf{X}:=\sum_{n=0}^{\infty}\mathbf{Y}^n$
and
\begin{equation*}
  x_n
  :=
  \left(
    y_1,\ldots,y_{n-1}
  \right),
\end{equation*}
it becomes impossible to compete even with the simplest prediction rules $D:\mathbf{X}\to\mathbf{Y}$:
there needs be no connection between the restrictions of $D$ to $\mathbf{Y}^n$
for different $n$.
The requirement that $y_1,\ldots,y_{n-1}$ should be compressed into an element $x_n$
of a locally compact space $\mathbf{X}$ restricts the set of possible prediction rules
so that it becomes manageable.
We can consider $\mathbf{X}$ to be the necessary bottleneck
in our notion of a prediction rule,
and the requirement of local compactness of $\mathbf{X}$ makes it narrow enough
for us to be able to compete with all continuous prediction rules.
A natural question is:
can the requirement of the local compactness of $\mathbf{X}$ be weakened
while preserving the existence of on-line prediction algorithms
competitive with the continuous prediction rules?
(And it should be remembered that our (\ref{eq:universal-consistency})
might be a poor formalization of the latter property
if sizeable pieces of $\mathbf{X}$ cannot be expected to be compact.)

\subsection*{Randomization}

It appears that various aspects of randomization
in this paper and competitive on-line prediction in general
deserve further study.
For example,
the bound of Corollary \ref{cor:decision-bounds}
is based on the worst possible outcome of Predictor's randomization
and the best possible outcome of the prediction rule's randomization
(disregarding an event of probability at most $\delta$).
This is unfair to Predictor.
Of course, comparing the expected values of Predictor's and the prediction rule's loss
would be an even worse solution:
this would ignore the magnitude of the likely deviations of the loss from its expected value.
It would be too crude to use the variance as the only indicator of the likely deviations,
and it appears that the right formalization should involve
the overall distribution of the deviations.

A related observation is that,
when using a prediction strategy based on defensive forecasting,
Predictor needs randomization only when
there are several very different predictions
with similar expected losses
with respect to the current probability forecast $P_n$.
Since $P_n$ are guaranteed to agree with reality,
we would not expect that Predictor will often find himself
in such a position
provided Reality is neutral
(rather than an active opponent).
Predictor's strategy will be almost deterministic.
It would be interesting to formalize this intuition.

\subsection*{Limitations of competitive on-line prediction}

In conclusion,
I will briefly discuss two serious limitations of this paper.

First, the main results of this paper only concern one-step-ahead prediction.
In a more general framework
the loss function would depend not only on $y_n$
but on other future outcomes as well.
There are simple ways of extending our results in this direction:
e.g.,
if the loss function $\lambda=\lambda(x_n,\gamma_n,y_n,y_{n+1})$
depends on both $y_n$ and $y_{n+1}$,
we could run two on-line prediction algorithms with the observation space $\mathbf{Y}^2$,
one responsible for choosing $\gamma_n$ for odd $n$ and the other for even $n$.
However, cleaner and more principled approaches are needed.

As we noted earlier (see Remark \ref{rem:Cover}),
the general interpretation of D-predictions
is that they are decisions made by a small decision maker.
To see why the decision maker is assumed small,
let us consider (\ref{eq:typical}),
which the kind of guarantee
(such as (\ref{eq:decision-bounds}))
provided in competitive on-line prediction
(although see \cite{cesabianchi/lugosi:2006}, Section 7.11,
for a recent advance).
Predictor's and the prediction rule $D$'s losses
are compared on the same sequence $x_1,y_1,x_2,y_2,\ldots$ of data and observations.
If Predictor is a big decision maker
(i.e., his decisions affect Reality's future behavior)
the interpretation of  (\ref{eq:typical})
becomes problematic:
presumably,
$x_1,y_1,x_2,y_2,\ldots$ resulted from Predictor's decisions $\gamma_n$,
and $D$'s loss should be evaluated on a different sequence:
the sequence $x^*_1,y^*_1,x^*_2,y^*_2,\ldots$
resulting from $D$'s decisions $D(x_n)$.

The approach of this paper is based on defensive forecasting:
the ability to produce ideal, in important aspects,
probability forecasts.
It is interesting that ideal probability forecasts are not sufficient in big decision making.
As a simple example, consider the game where there is no $\mathbf{X}$,
$\Gamma=\mathbf{Y}=\{0,1\}$,
and the loss function $\lambda$ is given by the matrix

\begin{center}
\begin{tabular}{c|cc}
             & $y=0$ & $y=1$ \\\hline
  $\gamma=0$ & 1     & 2 \\
  $\gamma=1$ & 2     & 0
\end{tabular}
\end{center}

\noindent
Reality's strategy is $y_n:=\gamma_n$,
but Predictor's initial theory is that Reality always chooses $y_n=0$.

Predictor's ``optimal'' strategy based on his initial beliefs
is to always choose $\gamma_n=0$ suffering loss $1$ at each step.
His initial beliefs are reinforced with every move by Reality.
Intuitively it is clear that Predictor's mistake in not choosing $\gamma_n\equiv 1$
is that he was being greedy
(concentrated on exploitation and completely neglected exploration).
However,
\begin{itemize}
\item
  he acted optimally given his beliefs,
\item
  his beliefs have been verified by what actually happened.
\end{itemize}
In big decision making
we have to worry about
what would have happened if we had acted in a different way.

My hope is that game-theoretic probability has an important role to play
in big decision making as well.
A standard picture in the philosophy of science
(see, e.g., \cite{popper:1934,kuhn:1962})
is that science progresses via struggle between (probabilistic) theories,
and it is conceivable that something like this also happens in individual (human and animal) learning.
Based on good theories
(the ones that survives serious attempts to overthrow them)
we can make good decisions.
Testing of probabilistic theories is crucial in this process,
and the game-theoretic version of the testing process
(gambling against the theory)
is much more flexible than the standard approach to testing statistical hypotheses:
at each time we know to what degree the theory has been falsified.
It is important, however, that the skeptic testing the theory
should not only do this playing the imaginary game with the imaginary capital;
he should also venture in the real world.
Predictor's theory that Reality always chooses $y_n=0$
would not survive for more than one round
had it been tested
(by choosing a sub-optimal, from the point of view of the old theory, decision).

Big decision making is a worthy goal
but it is very difficult to prove anything about it,
and elegant mathematical results might be beyond our reach
for some time.
Small decision making is also important
but much easier;
in many cases we can do it almost perfectly.

\subsection*{Acknowledgments}

I am grateful to the COLT'2006 co-chairs for inviting me to give the talk
on which this paper is based.
Theorems \ref{thm:decision-asymptotic} and \ref{cor:decision-asymptotic}
provide a partial answer to a question asked by Nicol\`o Cesa-Bianchi.
This work was partially supported by MRC (grant S505/65).

\end{document}